%% file: Formatting-Instructions-LaTeX-2025_new.tex
\documentclass[letterpaper]{article} 
\usepackage{aaai25}  
\usepackage{times}  
\usepackage{helvet}  
\usepackage{courier}  
\usepackage[hyphens]{url}  
\usepackage{graphicx} 
\usepackage{natbib}  
\usepackage{caption} 
\usepackage{algorithmic}
\usepackage[ruled,lined,linesnumbered]{algorithm2e}
\usepackage{float}
\usepackage{amssymb}
\usepackage{amsmath}
\usepackage{color}
\usepackage{xcolor}
\usepackage{newfloat}
\usepackage{listings}
\DeclareCaptionStyle{ruled}{labelfont=normalfont,labelsep=colon,strut=off} 
\floatstyle{ruled}
\newfloat{listing}{tb}{lst}{}
\floatname{listing}{Listing}
\title{MetaNeRV: Meta Neural Representations for Videos \\with Spatial-Temporal Guidance}
\author{
    Jialong Guo\textsuperscript{\rm 1}\equalcontrib,
    Ke Liu\textsuperscript{\rm 1}\equalcontrib,
    Jiangchao Yao\textsuperscript{\rm 2},
    Zhihua Wang\textsuperscript{\rm 1},
    Jiajun Bu\textsuperscript{\rm 1},
    Haishuai Wang\textsuperscript{\rm 1}\thanks{Corresponding author: Haishuai Wang}
}
\affiliations{
    \textsuperscript{\rm 1}Zhejiang Key Laboratory of Accessible Perception and Intelligent Systems, \\College of Computer Science,	Zhejiang University, China\\
    \textsuperscript{\rm 2}Cooperative Medianet Innovation Center, Shanghai Jiaotong University, China\\

    \{jialongguo, keliu99, zhihua\_wang, bjj, haishuai.wang\}@zju.edu.cn, 
    Sunarker@sjtu.edu.cn
}

\usepackage{bibentry}

\begin{document}

\maketitle

\begin{abstract}
Neural Representations for Videos (NeRV) has emerged as a promising implicit neural representation (INR) approach for video analysis, which represents videos as neural networks with frame indexes as inputs.  
However, NeRV-based methods are time-consuming when adapting to a large number of diverse videos, as each video requires a separate NeRV model to be trained from scratch. In addition, NeRV-based methods spatially require generating a high-dimension signal (i.e., an entire image) from the input of a low-dimension timestamp, and a video typically consists of tens of frames temporally that have a minor change between adjacent frames.
To improve the efficiency of video representation, we propose Meta Neural Representations for Videos, named MetaNeRV, a novel framework for fast NeRV representation for unseen videos. 
MetaNeRV leverages a meta-learning framework to learn an optimal parameter initialization, which serves as a good starting point for adapting to new videos.
To address the unique spatial and temporal characteristics of video modality, we further introduce spatial-temporal guidance to improve the representation capabilities of MetaNeRV. Specifically, the spatial guidance with a multi-resolution loss aims to capture the information from different resolution stages, and the temporal guidance with an effective progressive learning strategy could gradually refine the number of fitted frames during the meta-learning process. Extensive experiments conducted on multiple datasets demonstrate the superiority of MetaNeRV for video representations and video compression.
\end{abstract}

%
\begin{links}
    \link{code}{https://github.com/jialong2023/MetaNeRV}
\end{links}

\begin{figure}[!t]
\centering
\includegraphics[scale=0.43]{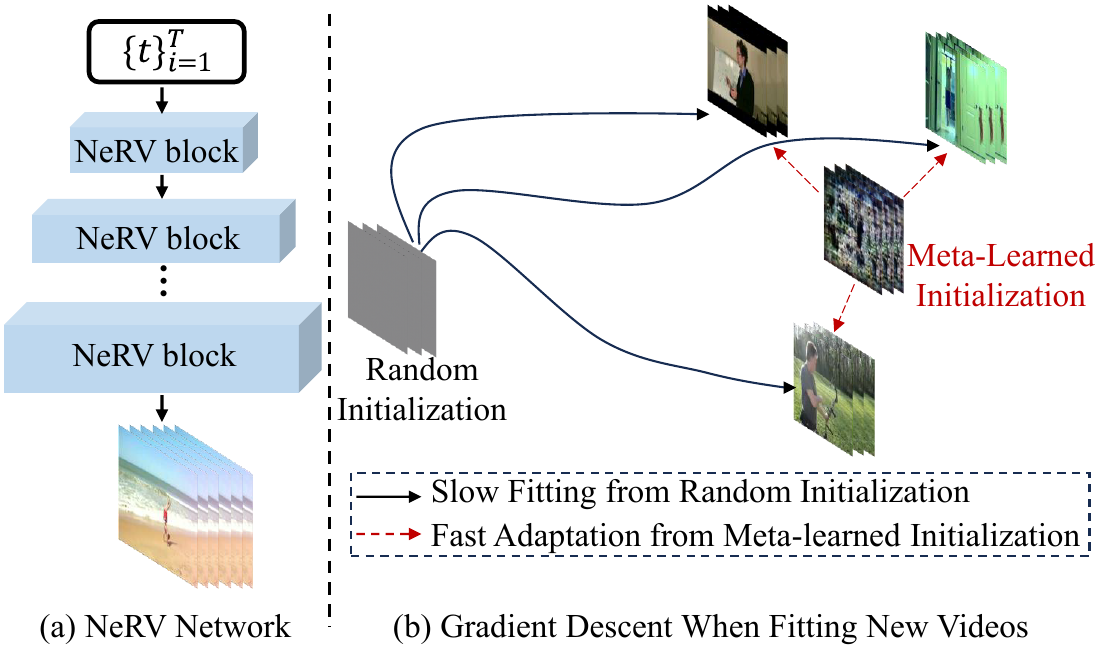}
\caption{(a) NeRV Network takes the frame index as input and outputs an image of that index. Querying a sequence of frame indexes results in a list of sequences, which can represent a video. (b) The network with random initialization necessitates optimization through numerous steps for new videos, whereas meta-learned initialization enables swift adaptation to new videos.}
\label{fig:frame1}
\end{figure}

\begin{figure*}[!t]
\centering
\includegraphics[scale=0.24]{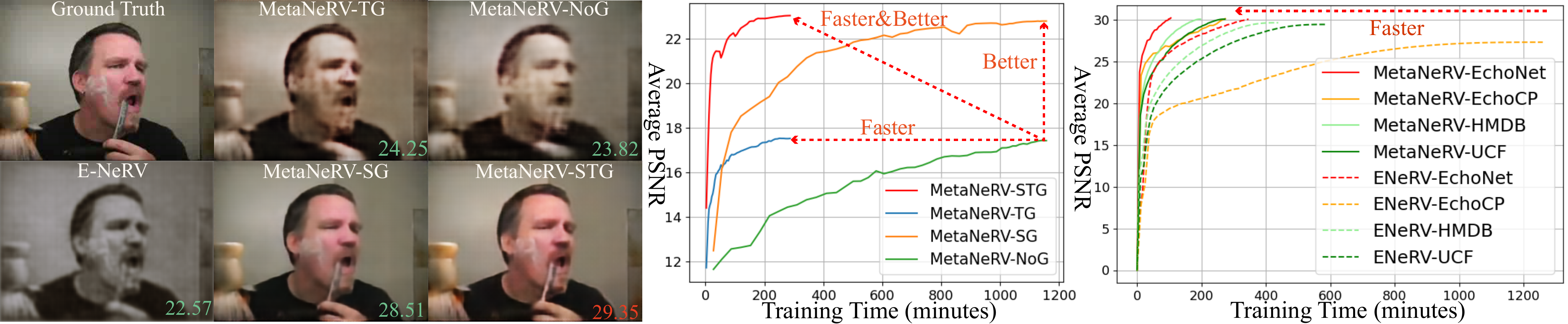}
\caption{(left) The visualization results with three-step inference in video representation tasks between the E-NeRV method and our method of the different guidance, where TG, SG, NoG, and STG respectively represent Temporal Guidance, Spatial Guidance, No Guidance, and Spatio-Temporal Guidance. (middle) The average PSNR and training time curves on the UCF dataset under different guidance, where the model trains faster and performs better under spatio-temporal guidance. (right) The average PSNR and training time curves of E-NeRV and our method on four datasets, given a target PSNR value of 30.}
\label{fig:frame_com}
\end{figure*}

\section{Introduction}
In recent years, Implicit Neural Representations (INR) have emerged as a powerful tool for continuously representing data in various computer vision tasks. The core idea of INR is to represent a signal as a function that can be effectively approximated by a neural network~\cite{pinkus1999approximation}. This network encodes the signal's values implicitly within its structure and parameters during the training or fitting process, allowing for the subsequent retrieval of these values through corresponding coordinates. Notably, INR has recently gained significant attention in the context of neural representations for videos, with notable examples including NeRV~\cite{chen2021nerv} and E-NeRV~\cite{li2022nerv}. In contrast to traditional coordinate-based neural representations, NeRV-based approaches take the frame index as input and directly generate the desired frame image, which is called image-wise implicit neural representations~\ref{fig:frame1}, resulting in significantly faster training and inference speeds compared to their coordinate-based counterparts.

Although existing NeRV-based methods demonstrate impressive capabilities, their limitation to encoding a single video at a time restricts their applicability in real-world scenarios, as each new video typically requires optimization through numerous gradient descent steps.

D-NeRV~\cite{he2023towards} memorize keyframes of videos within the training set, aiming to enhance generalization by reconstructing transition frames from these keyframes. Nevertheless, such methodologies are constrained to representing frames within the trained videos, potentially hindering their ability to generalize to previously unseen videos.

To accelerate the adaptation of NeRV-based models on unseen videos, we present \textbf{MetaNeRV}, a meta-learning framework designed to learn optimal initial weights for neural representations of videos. Compared to traditional random initialization techniques, MetaNeRV learns effective initialization weights across a series of videos, acting as a powerful prior that accelerates convergence during optimization and enhances generalization capabilities for unseen videos. 
Our methodology employs the optimization-based meta-learning algorithm MAML~\cite{finn2017model}, using a diverse meta-training set of videos to generate initial weight configurations ideally suited for representing unseen videos, leading to faster convergence and better generalization.
As depicted in Fig.~\ref{fig:frame1}(b), each new video necessitates its optimization process, which can be less efficient. However, with proper initialization, the number of iteration steps can be significantly reduced.


Relying solely on meta-learning may yield poor performance. To address this, we propose two enhancements:
Spatial Guidance: We introduce a multi-resolution loss function and add header modules to each NeRV block layer. These modules output video frames at different resolutions, improving the meta-learning framework's representation.
Temporal Guidance: To handle convergence issues and inefficiency with complex videos, we adopt a progressive training strategy. During meta-learning, we gradually increase subtask difficulty, allowing the framework to smoothly transition from easier to more complex tasks.

Experimental results on multiple datasets demonstrate that MetaNeRV outperforms other frame-wise methods in both video representations. 
Additionally, we explore various applications of our method, including video compression and video denoising tasks. 
With quantization-aware training and entropy coding, MetaNeRV outperforms widely-used video codecs such as H.264~\cite{wiegand2003overview} and HEVC~\cite{sullivan2012overview} and performs comparably with state-of-the-art video compression algorithms.

The contributions of this work are summarized as follows:

\begin{itemize}
    \item We introduce \textbf{MetaNeRV}, a meta-learning-driven framework with spatial-temporal guidance, tailored for NeRV-based video reconstruction. MetaNeRV enhances its performance by optimizing the initialization parameters.
    \item We propose a progressive training strategy as temporal guidance and incorporate a multi-resolution loss as spatial guidance within the meta-learning framework, providing precise supervision for better representation capabilities and improving training efficiency.
    \item Comprehensive experiments on various video datasets show that: (i) optimized initialization parameters lead to a significant acceleration in the convergence of the NeRV-based model, achieving a remarkable \textbf{9x} increase in speed; (ii) the incorporation of our proposed guidance enhances the training efficacy of the meta-learning framework, e.g. resulting in a notable improvement of \textbf{+16} PSNR in a single step on the EchoNet-LVH dataset; (iii) excellent performance in several video-related applications, including video compression and denoising.
\end{itemize}

\begin{figure*}[!t]
\centering
\includegraphics[scale=0.51]{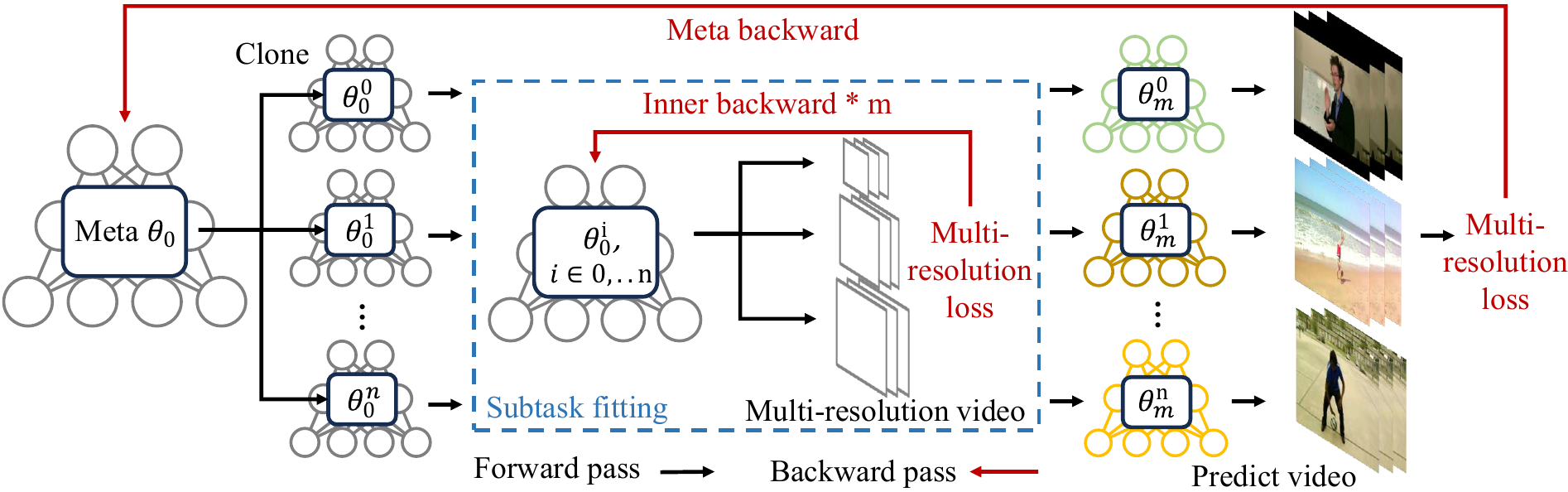}
\caption{Framework for MetaNeRV. A meta-learner is utilized to sample tasks of learning video and learns an initialized weight that can quickly fine-tune to a new video. The initialized weights will be cloned and then optimized m steps for n subtask in their corresponding video.}
\label{fig:frame2}
\end{figure*}

\section{Related Work}
\textbf{Implicit neural representations.}
Neural networks can be used to approximate the functions that map the input coordinates to various types of signals. It has brought great interest and has been widely adopted to represent 3D shape ~\cite{sitzmann2019scene, mescheder2019occupancy, park2019deepsdf, liu2023implicit}, novel view synthesis ~\cite{mildenhall2021nerf, 
yu2021pixelnerf} and so on. These approaches train a neural network to fit a single scene or object such that the network weights encode it. Implicit neural representations have also been applied to represent signals~\cite{hinton2006fast, vaswani2017attention, tancik2020fourier, sitzmann2020implicit, peng2022distill, peng2022hybridvocab, anr}, images~\cite{wang2003multiscale, hsu2019multi, chen2021learning, dupont2021coin, chen2024context, liu2023partition}, videos ~\cite{chen2021nerv, lai2021video, rho2022neural, he2022masked, tong2022videomae}, and time series~\cite{li2024tsinr}. 

\textbf{Image-wise implicit neural representations.}
The first image-wise implicit neural representation for videos is proposed by NeRV~\cite{chen2021nerv}, which takes the frame index and outputs the corresponding RGB frame. Compared to the pixel-wise implicit neural representation~\cite{sitzmann2020implicit}, NeRV improves the encoding and decoding speed greatly and achieves better video reconstruction quality. Based on NeRV, E-NeRV~\cite{li2022nerv} boosts the video reconstruction performance via decomposing the image-wise implicit neural representation into separate spatial and temporal contexts. 
CNeRV~\cite{chen2022cnerv} proposes a hybrid video neural representation with content-adaptive embedding to introduce internal generalization further. NRFF~\cite{he2023towards} introduces a visual content encoder to encode the clip-specific visual content from the sampled key-frames and a motion-aware decoder to output video frames. 
FFNeRV~\cite{lee2023ffnerv} introduces the multi-resolution temporal grids to combine different temporal resolutions.
HNeRV~\cite{chen2023hnerv} and HiVeRV~\cite{kwan2024hinerv} proposed a hybrid neural representation approach, employing a VAE-shaped deep network to address these concerns.

\textbf{Video compresion}
Visual data compression, a cornerstone of computer vision and image processing, has been extensively studied over several decades. Traditional video compression algorithms like H.264~\cite{wiegand2003overview}, and HEVC~\cite{sullivan2012overview} have achieved remarkable success. Some works have approached video compression as an image interpolation problem, introducing competitive interpolation networks~\cite{wu2018video}, generalized optical flow to scale-space flow for enhanced uncertainty modeling~\cite{agustsson2020scale, yang2020hierarchical}, and employed temporal hierarchical structures with neural networks for various components~\cite{yang2020learning}. However, these methods are still constrained by the traditional compression pipeline. Alternatively, NeRV adopts the INR method, transforming video compression into model compression and demonstrating substantial potential. Given that videos are typically encoded once but decoded multiple times, INR methods like NeRV excel due to their high decoding efficiency and facilitate parallel decoding, contrasting with sequential decoding requirements in other video compression methods post key frame reconstruction.

\textbf{Meta-learning INRs.}
Meta-learning typically addresses the problem of “few-shot learning”, where some example tasks are used to train an algorithm that has a great generalization ability on new similar tasks.
Some previous works on meta-learning have focused on few-shot learning ~\cite{ravi2016optimization, mishra2017simple, patravali2021unsupervised, liu2023partition} and reinforcement learning ~\cite{finn2017model, sitzmann2020metasdf}, where a meta-learner allows fast adaptation for new observations and better generalization with few samples.
Optimization-based meta-learning algorithms such as Model-Agnostic Meta-Learning (MAML)~\cite{finn2017model, li2017meta, antoniou2018train, flennerhag2019meta, rajeswaran2019meta, hospedales2021meta, tancik2021learned, guo2024self, liu2023partition} are relevant to this work. Given a network architecture for performing a task, these methods use an outer loop of gradient-based learning to find a weight initialization that allows the network to optimize more efficiently for new tasks at test time. 
These methods assume the use of a standard gradient-based optimization method such as stochastic gradient descent or Adam~\cite{kingma2014adam} at test time. 
 Liu et al.~\cite{liu2023partition} propose partition methods for learning-to-learn INRs by meta-learning.

We are the first to utilize the meta-learning framework for image-wise implicit neural representation models, resulting in increased convergence speed and enhanced generalizability of video implicit neural representation methods. As illustrated in Fig.~\ref{fig:frame_com}, our method has shown significant performance in quantitative and qualitative experiments.

\section{Methods}

\subsection{Problem Formulation}

We aim to learn a prior over INR for video. As in \cite{chen2021nerv, li2022nerv}, 
the video can be viewed as a continuous function $f: T \rightarrow F$ defined in a bounded domain that $T \in \mathbb{R}, F \in \mathbb{R}^{H \times W \times 3}$. 
We define a video $V=\{F_n\}_{n=1}^N $, where $F_n$ is the frame $\in \mathbb{R}^{H \times W \times 3}$ and $N$ denotes totally $N$ frames. 
The NeRV-based model fits the video via a deep neural network which is represented by a neural representation 
$f_\theta: \mathbb{R} \rightarrow \mathbb{R}^{H \times W \times 3}$, where the input is a frame index $t\in \mathbb{R}$ and the output is the corresponding RGB image $F_n\in \mathbb{R}^{H \times W \times 3}$.  
Therefore, video encoding is done by fitting a neural network $f_\theta$ to a given video. We present the details in Fig.\ref{fig:frame1}(a).

In NeRV-based models, the input usually consists of embedding vectors generated by frame index encoding. Some methods incorporate complementary information to generate these embedding vectors. We follow E-NeRV\cite{li2022nerv} to integrate coordinate data to construct spatiotemporal embedding vectors. These embedding vectors are then fed into the generator, which is successively expanded by convolutional or anti-convolutional operations to predict the desired image size. The theoretical details concerning video fitting can be found in Appendix Section A.

\subsection{MetaNeRV Framework}
In this section, we provide a detailed introduction to our MetaNeRV framework, as illustrated in Fig.~\ref{fig:frame2}. 
Given a dataset of observations of videos $\{V\}$ from a specific distribution $\mathcal{D}$ (e.g., traditional videos or ultrasound videos), our objective is to find initial weights $\theta_0^*$ that minimize the final loss $L(\theta_m, V)$ when optimizing a network $f_\theta$ through $m$ optimization steps to represent a new video from the same distribution. Our target function is formulated as follows:
\begin{equation}
\theta_0^*=\arg \min _{\theta_0} E_{V \sim \mathcal{D}}\left[L\left(\theta_m\left(\theta_0, V \right), V\right)\right].
\end{equation}

We utilize MAML~\cite{finn2017model, liu2023partition} to learn the initial weights of the network so that it can be a good starting point for gradient descent in the new tasks. 

Given a video $V$, calculating the weight values $\theta_m(\theta_0, V)$ necessitates executing $m$ optimization steps, collectively termed as the inner loop. We encapsulate this inner loop with an outer loop of meta-learning to ascertain the initial weights $\theta_0$. In each iteration of the outer loop, we sample a video $V_j$ from $\mathcal{D}$ and apply the update rule:
\begin{equation}
(\theta_0)_{j+1}=(\theta_0)_j-\left.\eta \nabla_\theta L\left(\theta_m\left(\theta, V_j\right), V_j\right)\right|_{\theta=(\theta_0)_j},
\end{equation}
with meta-learning learning rate $\eta$. This update rule applies gradient descent to the loss on the weights $\theta_m\left(\theta, V_j\right)$ resulting from the inner loop optimization.

We adopt a combination of L1 and SSIM loss as our loss function for network optimization, which calculates the loss of overall pixel locations of the predicted image and the ground-truth image. Loss function details can be found in Appendix Section A.

\begin{figure}[!t]
\centering
\includegraphics[scale=0.38]{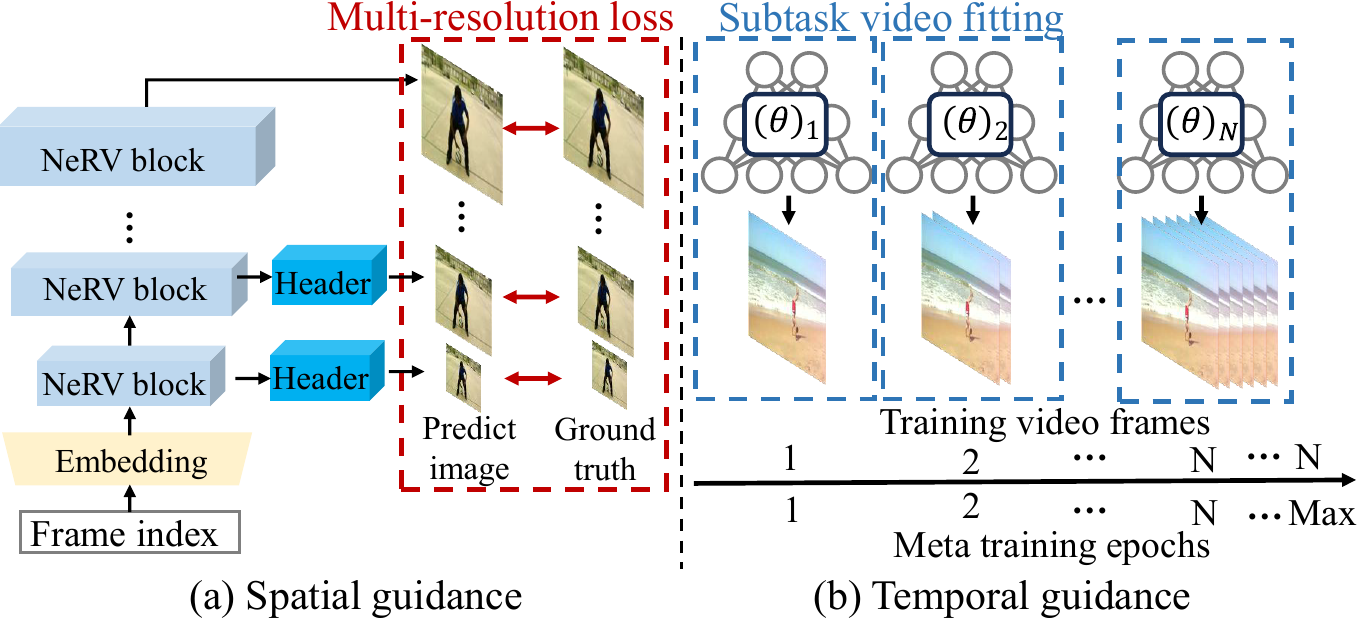}
\caption{
(a) NeRV network inputs a one-dimensional frame index, which expands through NeRV blocks to the image size, outputting corresponding frames. We propose adding a header block for spatial guidance at each NeRV block layer.
(b) We propose a progressive training strategy for temporal guidance, gradually increasing video frame numbers in subtasks during meta-learning.
}
\label{fig:frame3}
\end{figure}

\subsection{Spatial Guidance}
The spatial challenge arises because NeRV-based models progressively enlarge from a small vector by repeatedly passing through the same block. Directly applying the finest-grained supervision at the final resolution stages may result in insufficient supervisory signals for the preceding blocks, potentially making it difficult for the earlier resolution stages to converge to a better solution at that resolution. Therefore, we introduce multi-resolution supervision, which provides supervisory signals directly at different resolution stages, to encourage the output of all NeRV blocks to converge toward the unique ground truth. This spatial guidance enhances both the fitting accuracy and convergence speed.

The NeRV-based model employs up-sample blocks to scale the encoding of the frame index into the image with an appropriate block-by-block size. These up-sample blocks comprise $K$ feature layers, and for each of these layers, we append a convolutional header: 
\begin{equation}
\{F'_k\}_K=\{\texttt{header}_k(f_k)\}_K,
\end{equation}
where the feature of each layer $f_k$ is handled by the header layer $\texttt{header}_k$ to generate a video frame $F'_k$ corresponding to the size of the feature map as shown in Fig.~\ref{fig:frame3}(a).

This header transforms the multi-channel feature map into a three-channel feature map. We then downsample the ground truth image to align with the size of the feature map, ultimately computing the loss for gradient backpropagation:
\begin{equation}
L_{\rm multi}(V_j, \{F'_{k}\}_K)=\sum_{k=1}^K L(F'_k, \rm {Pooling}(V_j)),
\end{equation}
where the final loss $L_{\rm multi}$ is calculated by computing a weighted sum based on the global average pooling of downsample frame $X_j$.

\begin{table*}[]
\caption{The quantitative results of one-step and three-step inference for each method in five datasets.}
\label{tab:result1}
\resizebox{\linewidth}{!}{
\begin{tabular}{c|ccccc|ccccc}
\hline
Methods  & \multicolumn{5}{c|}{PSNR $\uparrow$(Step1/Step3)} & \multicolumn{5}{c}{MSSSIM $\uparrow$(Step1/Step3)} \\ 

 & MCL$\_$JCV & HMDB-51 & UCF101 & EchoNet-LVH & EchoCP & MCL$\_$JCV & HMDB-51 & UCF101 & EchoNet-LVH & EchoCP \\ \hline
NeRV         & 11.23/13.78 & 11.71/14.57 & 11.38/14.78 & 8.06/15.23  & 7.14/17.68 & 0.19/0.37 & 0.15/0.41 & 0.14/0.37  & 0.32/0.50 & 0.21/0.41 \\
E-NeRV            & 11.13/15.78 & 11.13/15.78 & 10.04/15.04 & 6.36/16.44  & 7.79/18.53 & 0.24/0.61 & 0.24/0.61 & 0.25/0.59  & 0.38/0.71 & 0.21/0.74 \\
FFNeRV              & 10.11/12.47 & 11.71/13.09 & 12.75/13.88 & 6.64/16.95 & 7.46/12.27 & 0.19/0.32 & 0.15/0.32 & 0.14/0.25 & 0.20/0.45 & 0.22/0.32 \\
HNeRV             & 11.25/12.69 & 11.71/12.89 & 10.9/13.89 & 6.72/17.35 & 6.61/17.57 & 0.21/0.36 & 0.15/0.3 & 0.13/0.26 & 0.23/0.53 & 0.20/0.40 \\
\textbf{MetaNeRV}    & \textbf{17.60/22.02} & \textbf{18.43/21.43} & \textbf{18.69/22.46} & \textbf{24.05/26.94} & \textbf{23.34/25.44} & \textbf{0.67/0.83} & \textbf{0.73/0.88} & \textbf{0.72/0.86} & \textbf{0.88/0.94} & \textbf{0.89/0.93} \\
\hline
\end{tabular}
}
\end{table*}

\begin{figure*}[!t]
\centering
\includegraphics[scale=0.23]{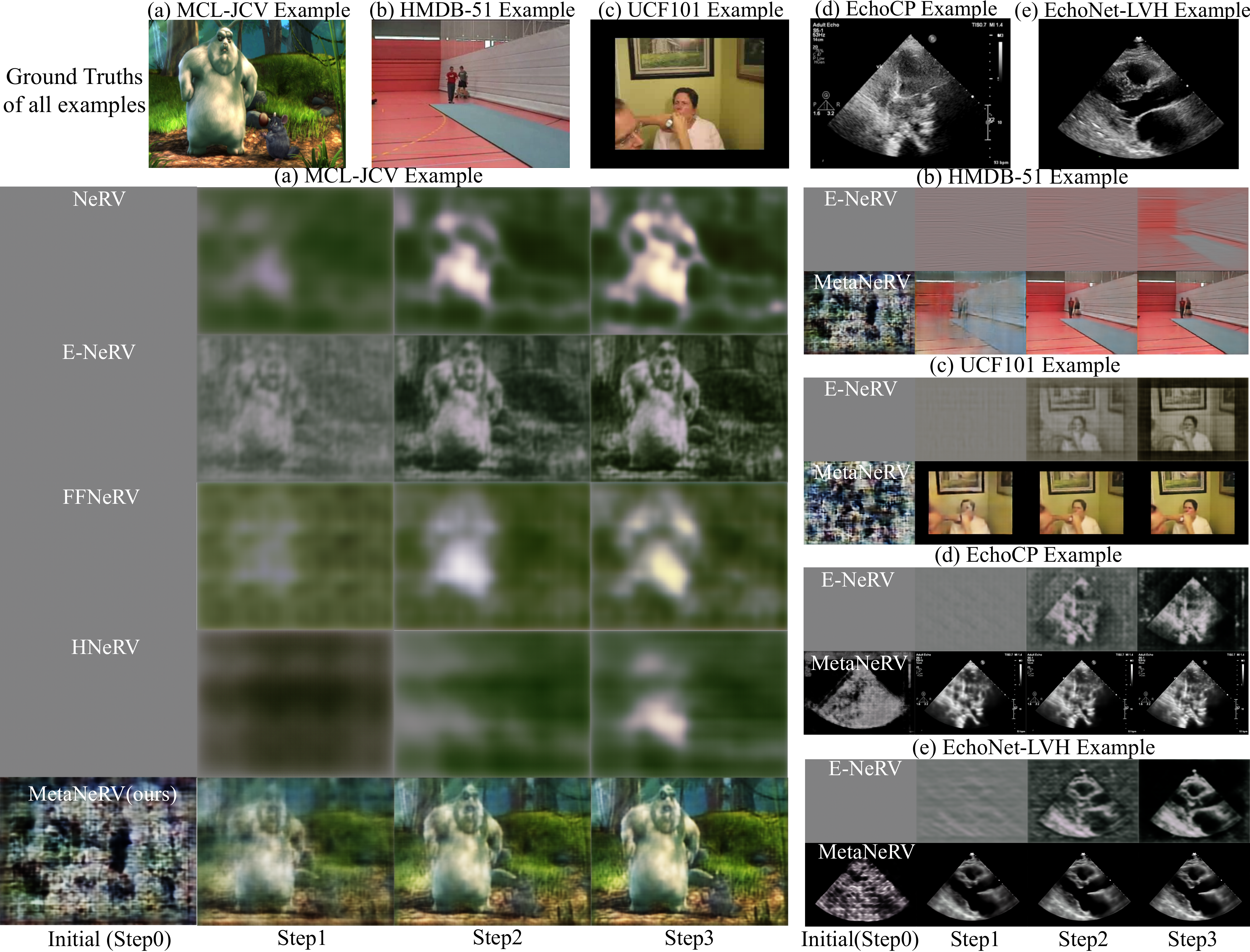}
\caption{
The visualization of NeRV, E-NeRV, FFNeRV, HNeRV, and MetaNeRV fitting the MCL$\_$JCV, HMDB-51, UCF101, EchoCP, and EchoNet-LVH examples. Notably, our method produces remarkable results in merely 3 iteration steps. 
``step 0" represents inference results directly from the initialization weight without further training.
}
\label{fig:mcl}
\end{figure*}

\subsection{Temporal Guidance}
In terms of time, a video typically consists of tens of frames, with minor changes between adjacent frames. For videos sharing similar backgrounds, utilizing spatial guidance during training can yield better results. However, when dealing with videos where background and foreground information exhibit significant differences, the training process encounters issues with low training efficiency.

To address this challenge and enhance the model's training efficiency, we introduce a progressive training strategy as temporal guidance. This method aims to assist the model in learning optimal initialization parameters from videos with distinct differences. As illustrated in Fig.~\ref{fig:frame3}(b), progressive training initiates the inner loop with a simple task: learning one frame per video. Subsequently, as iterations proceed, the number of frames in the videos is increased to elevate the task complexity gradually. The simple task enables the model to converge more rapidly, and by reducing the number of frames in the task, it also significantly decreases training time. Further details regarding the algorithm can be found in Appendix Section B.

\section{Experiments}

\subsection{Datasets and Implementation Details}
\label{sec:detail}

\textbf{Dataset} We conduct quantitative and qualitative comparison experiments on 8 different video datasets to evaluate our MetaNeRV against NeRV-based methods for video representation tasks. 
The datasets include multiple real-world datasets across various video types, such as UCF101\cite{soomro2012ucf101}, HMDB-51\cite{kuehne2011hmdb},  and MCL$\_$JCV~\cite{wang2016mcl}, as well as ultrasound datasets like EchoCP~\cite{wang2021echocp}, and EchoNet-LVH~\cite{duffy2022high}.
We selected 900 videos for each dataset, 800 for the training set, and 100 for the test set. 
Each video sequence contains 60 frames, processed to a resolution of 320×240.

Furthermore, we conduct inference experiments on HOLLYWOOD2~\cite{marszalek09}, SVW~\cite{safdarnejad2015sports}, and OOPS~\cite{epstein2020oops}, which are diverse datasets for action recognition, encompassing movie scenes, amateur sports, and unintentional human activities, with 300 videos in each dataset. A description of more datasets can be found in Appendix Section C.

\begin{figure}[!t]
\centering
\includegraphics[scale=0.29]{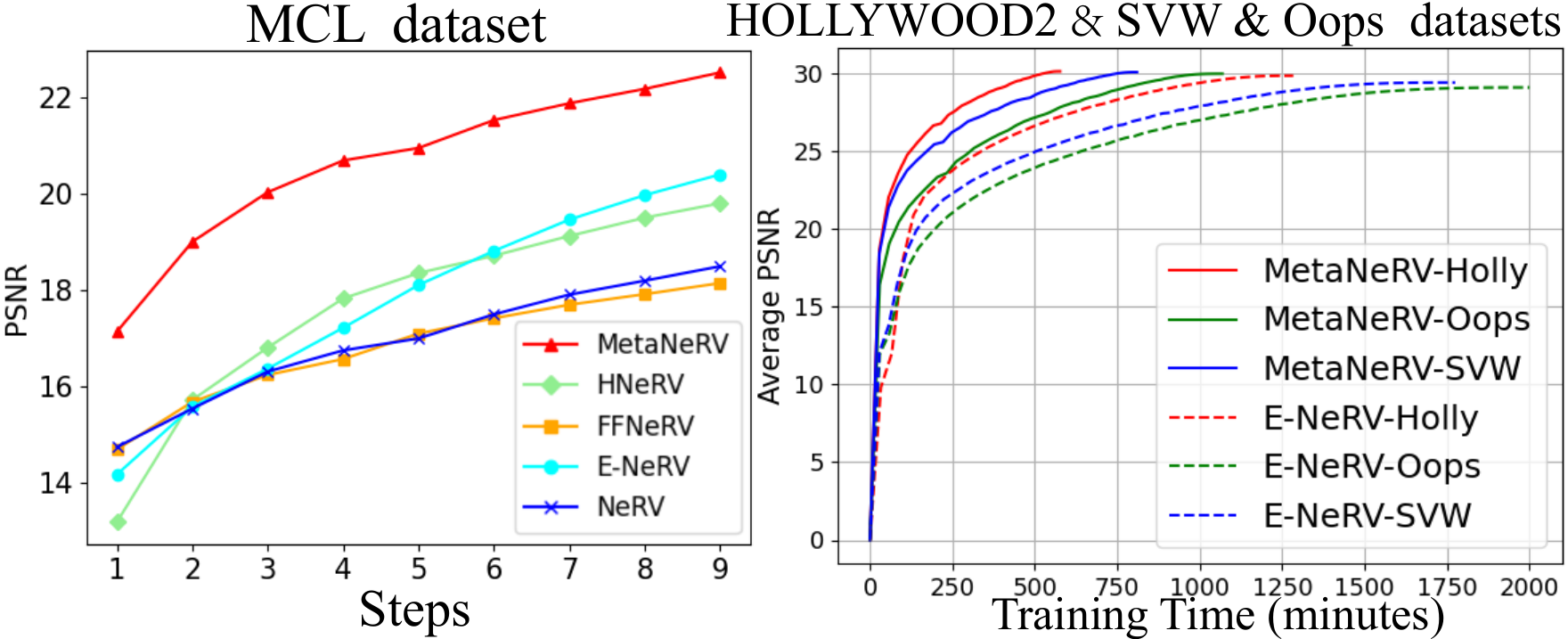}
\caption{
(left) Comparison of baselines, and our MetaNeRV on MCL$\_$JCV dataset. Our method’s performance on 1 step is better than FFNeRV’s at 9 steps, which shows better performance and faster convergence. (right) Comparison of MetaNeRV and E-NeRV on HOLLYWOOD2, SVW, and Oops datasets. Given a training objective of achieving an average PSNR of 30 for the dataset, our model significantly reduces training time.}
\label{fig:mcl_psnr}
\end{figure}

\begin{figure*}[!t]
\centering
\includegraphics[scale=0.18]{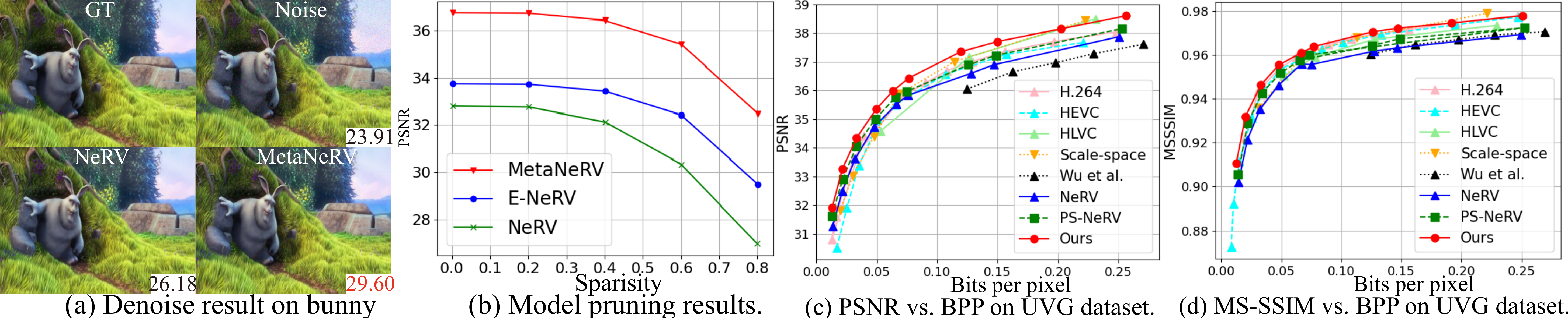}
\caption{
(a) Better denoise result on Bunny data. ``Noise" denotes the noisy frames before any denoising process. (b) MetaNeRV outperforms other methods in pruning. (c)(d) MetaNeRV shows better video compression results on the UVG dataset.
}
\label{fig:noise}
\end{figure*}

\textbf{Implementation} We set up-scale factors $5,2,2,2,2$ for each block of our MetaNeRV model to reconstruct a 320×240 image from the feature map of size 4×3. For a fair comparison, we follow the training schedule of the original E-NeRV implementation. We train the model using Adam optimizer~\cite{kingma2014adam} with a learning rate 1e-4 by Pytorch. We conduct all experiments with RTXA6000 GPU, 
while the number of inner loop steps is 3. 

\textbf{Metrics} For evaluation metrics, we use PSNR and MS-SSIM~\cite{wang2003multiscale} to evaluate reconstruction quality. 
Bits-per-pixel (BPP) is adopted to evaluate the image compression performance.

\subsection{Main Results}
\label{sec:main}

\subsubsection{Video representation.}
Initially, we compare our method with image-wise INR methods. Our model has trained optimal initialization weights separately on real-world datasets (HMDB-51, UCF101) and ultrasound datasets (EchoCP, EchoNet-LVH). Due to the limited number of videos in the MCL$\_$JCV dataset, which hinders the optimization of better initialization weights, we directly infer on MCL$\_$JCV using weights trained on HMDB-51.

\textbf{Qualitative comparison}. 
Our method can swiftly adapt to the content of new videos, even with just a few iteration steps in the video representation task, as shown in Fig.~\ref{fig:mcl}.
We observe that visualizations of initial weights on real-world datasets appear more reasonable than visualizations from randomly initialized weights, while those on ultrasound datasets exhibit more dataset-specific characteristics, visually demonstrating that our method has learned an optimal initialization weight.

\textbf{Quantitative comparison}. Our method significantly all other outperforms image-wise methods under fewer iteration steps. as presented in Tab.~\ref{tab:result1}. 
Notably, on ultrasound datasets, our method's PSNR and MS-SSIM metrics in one-step iteration exceed those of others by $300\%$. On real-world datasets, our method significantly outperforms other methods in both one-step and three-step iterations. 

\textbf{OOD results}. To demonstrate the generalization and efficiency of our method, we conducted extensive experiments on out-of-distribution (OOD) datasets. All experiments in Fig.~\ref{fig:mcl_psnr} used weights trained on HMDB-51 and directly inferred on four OOD datasets. The left side of Fig.~\ref{fig:mcl_psnr} showcases our method's excellent performance on MCL$\_$JCV, while the right side illustrates that, given a target PSNR value for training, our method can significantly reduce video representation time and improve efficiency on three datasets with a larger number of videos.

\begin{table*}[]
\caption{The ablation quantitative results of one-step and three-step inference for each method in four datasets.}
\label{tab:result2}
\resizebox{\linewidth}{!}{
\begin{tabular}{cccc|cccc|cccc}
\hline
Methods & Meta- & Temporal & Spatial & \multicolumn{4}{c|}{PSNR $\uparrow$(Step1/Step3)} & \multicolumn{4}{c}{MSSSIM $\uparrow$(Step1/Step3)} \\ \cline{5-12}
& learning & Guidance & Guidance & HMDB-51 & UCF101 & EchoNet-LVH & EchoCP & HMDB-51 & UCF101 & EchoNet-LVH & EchoCP \\ \hline
E-NeRV   & $\times$  & $\times$     & $\times$           & 11.13/15.78 & 10.04/15.04 & 6.36/16.44  & 7.79/18.53 & 0.24/0.61 & 0.25/0.59  & 0.38/0.71 & 0.21/0.74 \\
MetaNeRV-NoG     & $\checkmark$ & $\times$     & $\times$           & 16.89/19.43 & 16.4/19.08 & 21.66/23.72 & 19.77/21.78 & 0.57/0.71 & 0.55/0.7 & 0.81/0.87 & 0.79/0.87 \\
MetaNeRV-TG   & $\checkmark$ & $\checkmark$  & $\times$           & 17.3/19.31 & 16.83/19.53 & 22.16/24.31 & 20.63/23.15 & 0.63/0.76 & 0.59/0.73 & 0.83/0.88 & 0.81/0.89 \\
MetaNeRV-SG   & $\checkmark$ & $\times$     & $\checkmark$        & 17.41/22.06 & 18.04/21.96 & 23.19/25.63 & 22.03/23.77 & 0.69/0.85 & 0.68/0.83 & 0.85/0.91 & 0.88/0.92 \\
\textbf{MetaNeRV-STG} & \checkmark & \checkmark  & \checkmark        & \textbf{18.43/21.43} & \textbf{18.69/22.46} & \textbf{24.05/26.94} & \textbf{23.34/25.44} & \textbf{0.73/0.88} & \textbf{0.72/0.86} & \textbf{0.88/0.94} & \textbf{0.89/0.93} \\
\hline
\end{tabular}
}
\end{table*}

\begin{figure*}[!t]
\centering
\includegraphics[scale=0.26]{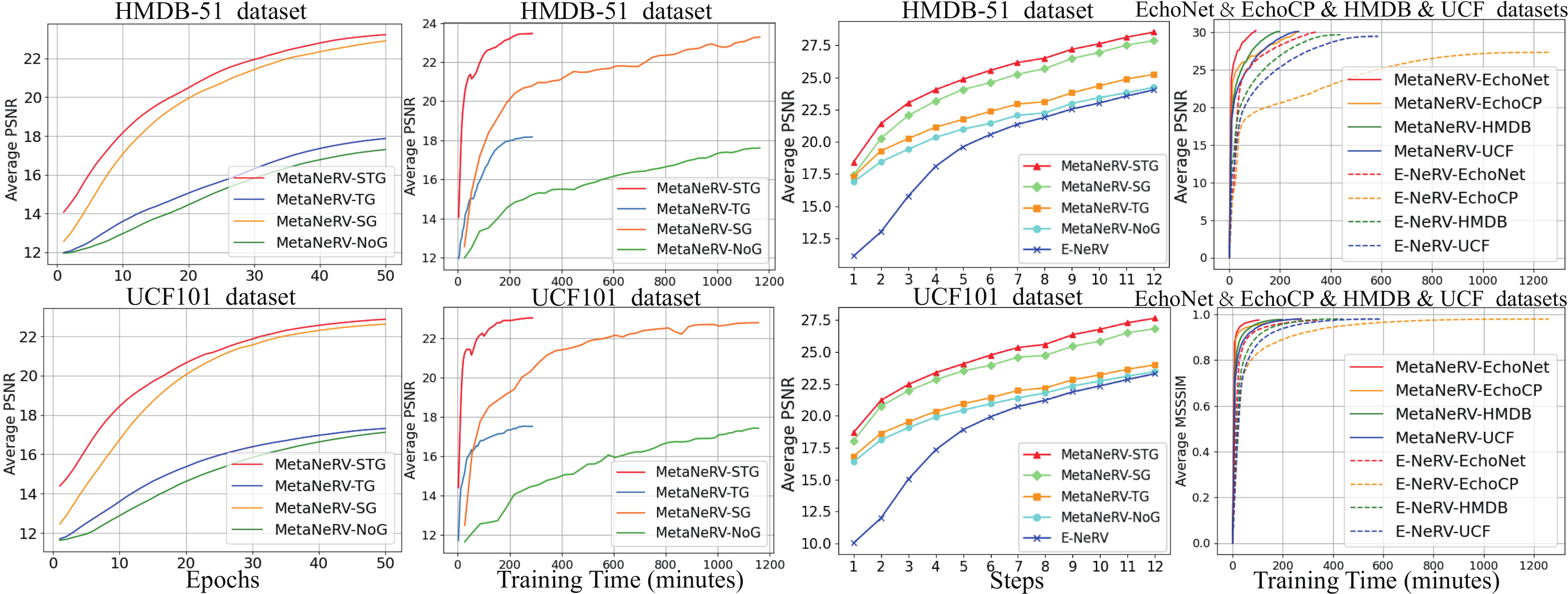}
\caption{
(a) Training curves of Epochs vs PSNR with different guidance, demonstrating improved model performance with spatial guidance.
(b) Training curves of Time vs PSNR with different guidance, proving reduced training time and enhanced efficiency with temporal guidance.
(c) Inference performance of our method variants, all surpassing the baseline.
(d) Our method significantly reduces representation time given a 30 PSNR target.
}
\label{fig:psnr and ssim 1}
\end{figure*}

\subsubsection{Video Compression and Denoising.}
We further evaluate MetaNeRV's versatility with two downstream tasks: 1) video denoising on the Bunny data, and 2) video compression on the UVG dataset. 
Adhered to NeRV's setting, we apply videos with noise as training data, and compare the prediction results with the real videos.
We also apply an additional neural network parameter pruning with various prune ratios for different NeRV-based methods to evaluate the video compression performance. In addition, we compare the compression ability of our methods with lots of popular methods, including H.264\cite{wiegand2003overview}, HEVC\cite{sullivan2012overview}, HLVC\cite{yang2020learning}, Scale-space\cite{agustsson2020scale}, Wu et al.\cite{wu2018video}, NeRV\cite{chen2021nerv}, and PS-NeRV\cite{bai2023ps}.

\textbf{Stronger denoising ability}. As shown in Fig.~\ref{fig:noise}(a), we observe that the denoising result from MetaNeRV achieves better visualization performance and higher PSNR than the denoising result from NeRV, which demonstrates the strong denoising ability of MetaNeRV.

\textbf{Better performance for network pruning}. As depicted in Fig.~\ref{fig:noise}(b), MetaNeRV achieves better reconstruction PSNR at all different sparsity (pruning with different parameter ratios) than NeRV and E-NeRV, highlighting its robust network pruning ability for video compression.

\textbf{More powerful compression ability}. 
We present the rate-distortion curves in Fig.\ref{fig:noise}(c). We find that MetaNeRV surpasses all image-wise NeRV-based approaches. Furthermore, MetaNeRV outperforms traditional video compression technologies and other learning-based video compression methods at most BPPs.

\subsection{Ablation Studies}
We also conducted ablation studies on four datasets to verify the effect of different guidance. More qualitative and quantitative results can be found in Appendix Section D.

\textbf{Meta-learning}. 
The meta-learning framework accelerates the video representation of four datasets by gradually adding both guidances across all experiment video datasets, as presented in Tab.\ref{tab:result2}.

\textbf{Spatial guidance}. 
The spatial guidance significantly enhances the model's performance as shown in Fig.\ref{fig:psnr and ssim 1} (a) of training curves. 
Furthermore, during the inference stage in Fig.\ref{fig:psnr and ssim 1} (c), the method with added spatial guidance achieves a higher PSNR under the same number of iterations, indicating its effectiveness in improving the model's performance.

\textbf{Temporal guidance}. 
The temporal guidance is effective in reducing the model's training time while also achieving a slight performance improvement, as shown in Fig.\ref{fig:psnr and ssim 1} of training curves. During the inference stage, the model exhibits good performance, indicating that our proposed temporal guidance enhances training efficiency without compromising the model's overall performance.

\textbf{Significant results}.
Remarkably, our proposed method shows outstanding results after one iteration, surpassing the baseline by over 16 PSNR and 3x MS-SSIM on EchoNet-LVH, and consistently exceeding by at least 4 PSNR across other datasets.
Fig.\ref{fig:psnr and ssim 1} (d) illustrates that our proposed method significantly enhances the efficiency of video representation under the given training objectives.

\section{Conclusion}
In conclusion, we present MetaNeRV, a sophisticated meta-learning framework designed to optimize the initialization process of NeRV models. By learning optimal initialization parameters, we have achieved significant improvements in both the performance and efficiency of video reconstruction tasks. We introduce spatial guidance for precise training supervision and a temporal guidance regimen to enhance training efficiency while maintaining stability. 
The model has limitations, it cannot represent video for higher resolutions without retraining and may face convergence challenges with limited video training data.
Future work will focus on addressing these constraints.

\section{Acknowledgements}

This work is supported by the National Key R\&D Program of China (Grant No. 2022ZD0160703), the National Natural Science Foundation of China (Grant Nos. 62202422, 62406279, and 62071330), and the Shanghai Artificial Intelligence Laboratory.

\bibliography{aaai25}

\clearpage
\appendix
\input{appendix}

\end{document}

%% file: appendix.tex
\section{Appendix}
\subsection{A. Video Fitting Details}
\label{sec:fit}
In implicit neural representation for video, the video $V$ is estimated and parameterized by a neural function $f_\theta$ with $\theta$ as its weights (learnable parameters). 
A typical example of $f_\theta$ is a NeRV~\cite{chen2021nerv}. We consider a more general class of $f_\theta$ where the model fits video by mapping the inputs to high embedding space and upscaling by the MLP layers. Specifically, the basic NeRV uses Positional Encoding ~\cite{tancik2020fourier} as its emz2bedding function, while the E-NeRV separately embeds the spatial-temporal context and is fused by the transformer ~\cite{vaswani2017attention}.

\begin{equation}
\mathbf{e}=F_\mathbf{f}(F(\gamma(\mathbf{t}))),
\end{equation}
where $\gamma$ is frequency positional encoding, and $F$ is an MLP network encoder of the timestamp into a vector. Here 
$F_\mathbf{f}$ stands for a network that fuses the other information into the temporal embeddings like spatial information. $\mathbf{e}$ stands for the 
final embedding of the temporal or spatial-temporal information, and the input of the next upscaleed block.

\begin{figure}[!t]
\centering
\includegraphics[scale=0.44]{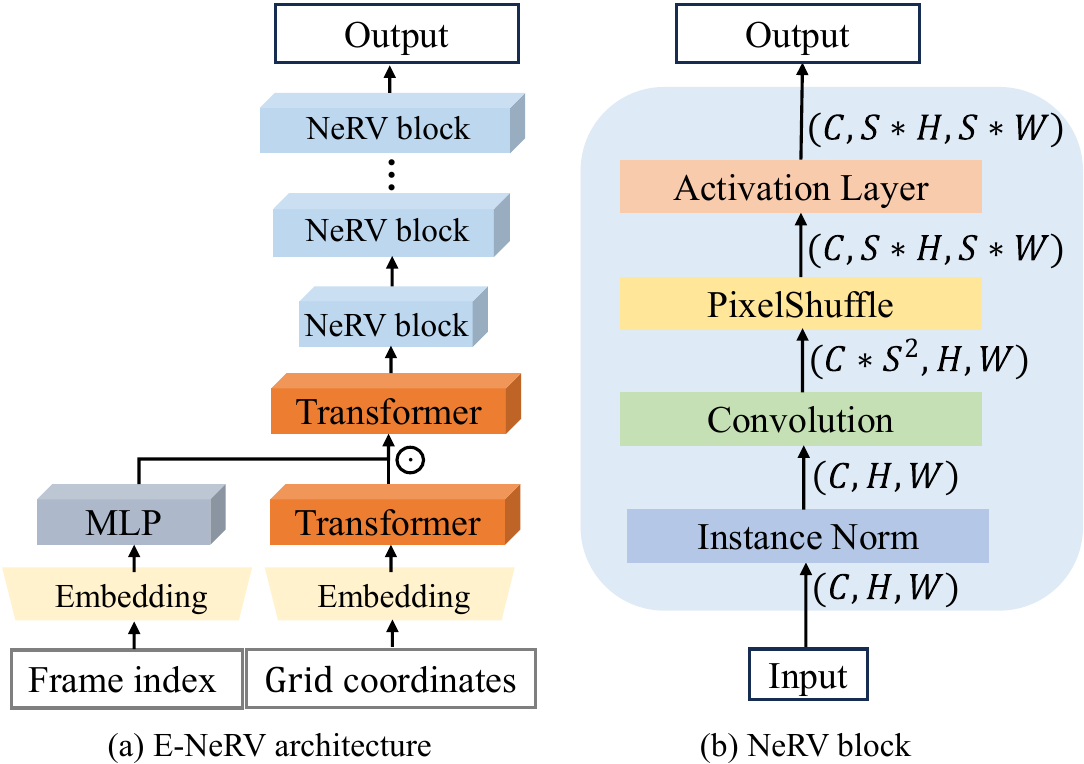}
\caption{
(a) Architecture of E-NeRV.  (b) NeRV block.
}
\label{fig:framework5}
\end{figure}

The NeRV-based model uses MLP blocks as upscaled blocks, and the final embedding is upscaled to the output image size by stacking multiple layers of upscaled blocks. Suppose we have L layers, each of which performs an upscale operation. For layer $(l) ((1 \leq l \leq L))$, the upscaling operation can be expressed as:

\begin{equation}
f_{l} = \text{Upscale}(f_{l-1}, W_l, b_l),
\end{equation}
where $(f_{l-1})$ is the output of the previous layer (for the first layer, the temporal embedding vector $\mathbf{e}$) and $W_l$ and $b_l$ are the weight and bias of layer $(l)$. $(\text{Upscale})$ can be any appropriate upscaling function, such as convolution, fully connected layers, etc. Here the NeRV block is shown in Fig.~\ref{fig:framework5}.

In each layer, feature transformations and nonlinear activations can be performed in addition to upscaling operations. This can be expressed as:

\begin{equation}
f_{l} = \sigma(W_l \cdot f_{l-1} + b_l),
\end{equation}
where $(\sigma)$ is a nonlinear activation function such as ReLU~\cite{glorot2011deep} and GeLU~\cite{hendrycks2016gaussian}.

After the (L) layer of processing, we obtain a representation $(f_L)$ that matches the image aspect size. This representation can be considered as a feature map where each layer contains image information related to the input time (t).

In summary, the whole process can be formulated as:

\begin{equation}
f_L = \text{NeRV}(t, \{W_l, b_l\}_{l=1}^{L}, \sigma, \text{Upscale}),
\end{equation}
here, $(f_L)$ is the final output feature map, associated with the input time t and has the same spatial dimension as the target image. e is the temporal embedding function, $(W_l)$ and $(b_l)$ are the weight and bias of the layers,  $\sigma$ is the nonlinear activation function, and $(\text{Upscale})$ is the upscaling function.

We apply gradient descent to fit the neural network. Let $\theta_0$ denote the initial network weights before any gradient steps are taken, and let $\theta_i$ denote the weights after $i$ steps of optimization. Basic gradient descent is applied following the rule:
\begin{equation}
\theta_{i+1}=\theta_i-\left.\beta \nabla_\theta L(\theta)\right|_{\theta=\theta_i},
\end{equation}
where $\beta$ is a learning rate parameter with an optimizer, 
to track the gradient moments over time to redirect the optimization trajectory. 
Given $m$ optimization steps, different initial weights $\theta_0$ will lead to different final weights $\theta_m$ and loss $L(\theta_m)$.

We adopt a combination of L1 and SSIM loss as our loss function for network optimization, which calculates the loss of overall pixel locations of the predicted image and the ground-truth image. Loss function as follows:

\begin{flalign*}    
&L_{Fusion} = \sum_{n=1}^N \left\{L(\theta |f_\theta(t),F_n )\right\} & \\    
&=\sum_{n=1}^N \left\{\alpha\left\|f_\theta(t)-F_n\right\|_1+(1-\alpha)\left(1-\operatorname{SSIM}\left(f_\theta(t), F_n\right)\right)\right\} &    
\end{flalign*}
where $\alpha$ is the hyper-parameter to balance the weight for each loss component, and $F_n$ is the responding video frame of frame index $t$. The total number of video frames is $N$.

We adopt E-NeRV~\cite{li2022nerv} with $12.49M$ parameters as our baseline. We follow the same settings as in E-NeRV, like activation choice, as shown in Fig.~\ref{fig:framework5}. We set $d=d_t=196$ for spatial and temporal feature fusion and $d_0=128$ for temporal instance normalization. We set all the positional encoding layers in our model identical to E-NeRV’s positional encoding formulated, and we use $b=1.25$ and $l=80$ if not otherwise denoted. 

In this work, we implemented the proposed MetaNeRV and other NeRV-based models ~\cite{chen2021nerv, li2022nerv, chen2023hnerv, lee2023ffnerv} using the PyTorch framework. And we adopted their original implementations for training NeRV, E-NeRV, HNeRV, and FFNeRV.

\subsection{B. Algorithm Details}
In this section, we provide a detailed introduction to the application of spatial-temporal guidance in meta-learning methods. 
The algorithm~\ref{alg} details the progressive training approach with a NeRV-based model. 
The network parameters are initially set with predefined values, denoted as $\theta$. The training commences with the first frame $F_1$ and progressively incorporates subsequent frames until all $N$ frames have been utilized. 
In the outer-loop iteration $j$, the network receives the top $T=j$ frames from $F_1$ to $F_T$ as input (if $j>N$, we choose $T=N$). 
The network computes the multi-resolution output $\{F'_{k}\}_K$ based on the current parameters $\theta$ for all $T$ frames. Subsequently, a multi-resolution loss function evaluates the difference between the actual video $X_j$ and the network's output $\{F'_{k}\}_K$, generating a loss value $L_{i,j}$, where $i$ indicates the $i^{th}$ inner loop result and $j$ indicates the $j^{th}$ outer loop. 

The algorithm updates the network parameters $\theta$ with the previous loss. This update is performed using gradient descent, specifically by calculating the average gradient across all losses incurred in iteration $t$ (i.e., $L_{1,j}, L_{2,j}, \ldots, L_{m,j}$). The learning rate $\eta$ regulates the step size in the parameter update. As the training progresses from iteration to iteration, more frames are incorporated, enabling the network to learn from a growing dataset. The updated equation from equation (3) can be represented as:
\begin{equation}
(\theta_0)_{j+1} = (\theta_0)_{j} - \eta \frac{1}{m} \sum_{i=1}^{m} \nabla_{\theta} L_{i,j}.
\end{equation}

\begin{algorithm}[!t]
\SetAlgoLined  
\KwIn{Distribution $\mathcal{D}$ over Video samples, outer learning rate $\eta$, number of inner-loop steps $m$, number of outer-loop steps $M$, number of feature layers $K$}  
\KwOut{Best initialized parameters $\theta^*$, best per-parameter inner-loop learning rates $\beta^*$}  
Initialize per-parameter inner-loop learning rates $\beta$ and network parameters with $\theta$\; 
\For {\textbf{outer loop }$j = 1$ \KwTo $M$}{
    Sample a video $V_j \sim \mathcal{D}$\;
    \textcolor{blue}{Get the top $T=j$ frames: $V_j\leftarrow V_j[:j]$}\; 
    Initialize $\phi_j = \theta$\;
    \For{\textbf{inner loop }$i = 1$ \KwTo $m$}{  
        \textcolor{red}{Predict network output with $K$ different resolution $\{F'_{k}\}_K = \text{NeVR-based Network}(\phi_j)$}\; 
        \textcolor{red}{Compute loss among all $T$ frames: $L_{i,j} = L_{\rm multi}(V_j, \{F'_{k}\}_K)$}\;      
    Update $\phi_j = \phi_j - \beta\nabla_{\phi_j}L_{i,j}$\;
    }  
    Update $\theta$ with $\{L_{1,j}, L_{2,j}, ..., L_{m,j}\}$: $\theta = \theta - \eta \frac{1}{m} \sum_{i=1}^{m} \nabla_{\theta} L_{i,j}$\;
    Update $\beta$ with $\{L_{1,j}, L_{2,j}, ..., L_{m,j}\}$: $\beta = \beta - \eta \frac{1}{m} \sum_{i=1}^{m} \nabla_{\beta} L_{i,j}$\;
    
}
\Return parameters $\theta$, inner-loop learning rate $\beta$\; 
\caption{MAML Training with Multi-resolution loss (in Red) and Progressive Training for NeRV (in Blue)}  
\label{alg}
\end{algorithm}

\subsection{C. Datasets Details}
\label{sec:data}
\textbf{UCF101}~\cite{soomro2012ucf101} dataset is an extension of UCF50 and consists of 13,320 video clips, which are classified into 101 categories. These 101 categories can be classified into 5 types (Body motion, Human-human interactions, Human-object interactions, Playing musical instruments, and Sports). The total length of these video clips is over 27 hours. All the videos are collected from YouTube and have a fixed frame rate of 25 FPS with a resolution of 320 × 240.

\textbf{EchoCP}~\cite{wang2021echocp} dataset is an echocardiography dataset in contrast to transthoracic echocardiography (cTTE) targeting PFO (Patent foramen ovale) diagnosis. EchoCP consists of 30 patients with both rest and Valsalva maneuver videos which cover various PFO grades. The video is captured in the apical-4-chamber view and contains at least ten cardiac cycles with a resolution of 640 × 480. 

\textbf{EchoNet-LVH}~\cite{duffy2022high} dataset is a standard echocardiogram study consisting of a series of videos and images visualizing the heart from different angles, positions, and image acquisition techniques. The EchoNet-LVH dataset contains 12,000 parasternal-long-axis echocardiography videos from individuals who underwent imaging as part of routine clinical care at Stanford Medicine. Each video was cropped and masked to remove text and information outside of the scanning sector. The resulting videos are at native resolution.

\textbf{HMDB-51} ~\cite{kuehne2011hmdb} dataset is a large collection of realistic videos from various sources, including movies and web videos. The dataset is composed of 6,766 video clips from 51 action categories (such as “jump”, “kiss” and “laugh”), with each category containing at least 101 clips. The original evaluation scheme uses three different training/testing splits. In each split, each action class has 70 clips for training and 30 clips for testing. The average accuracy over these three splits is used to measure the final performance.

\textbf{MCL-JCV} ~\cite{wang2016mcl} dataset consists of 24 source videos with resolution 1920×1080 and 51 H.264/AVC encoded clips for each source sequence. Single-pass constant QP encoding (CQP) was used with the Quantization Parameter (QP) ranging from 1 to 51. More than 120 volunteers participated in the subjective test. Each set of sequences was evaluated by around 50 subjects in a controlled environment.

\textbf{HOLLYWOOD2} \cite{marszalek09}:
This dataset contains 3669 video clips, totaling 20.1 hours, featuring 12 human action and 10 scene classes from 69 movies, providing a benchmark for real-world action recognition.

\textbf{SVW} \cite{safdarnejad2015sports}:
Containing 4200 videos from the Coach’s Eye app, SVW covers 30 sports categories and 44 actions. Its amateur content poses significant challenges for automated analysis.

\textbf{OOPS} \cite{epstein2020oops}:
The OOPS dataset includes 20,338 diverse YouTube videos totaling over 50 hours, capturing unintentional human actions in various environments and with differing intentions.

\begin{table}[]  
\resizebox{\linewidth}{!}{
\begin{tabular}{c|cc|cc}  
\hline  
 & \multicolumn{2}{c|}{Ultrasound} & \multicolumn{2}{c}{Real-world} \\ \hline
 Initialization & EchoNet & EchoCP & HMDB51 & UCF101 \\ \hline 
 Random  & 7.15/18.74 & 7.79/18.53 & 11.93/17.08 & 10.83/16.34 \\
 EchoNet & \textcolor{red}{24.05/26.94} & \textcolor{blue}{18.63}/19.41 & 10.62/15.11 & 12.42/17.62 \\
 EchoCP & \textcolor{blue}{23.49/26.60} & \textcolor{red}{23.34/25.44} & 11.03/15.68 & 12.46/17.97 \\ 
 HMDB51 & 13.72/22.82 & 16.89/21.19 & \textcolor{red}{18.43/23.02} & \textcolor{blue}{18.52/22.55} \\
 UCF101 & 14.93/24.58 & 17.71/\textcolor{blue}{21.59} & \textcolor{blue}{18.22/22.65} & \textcolor{red}{18.69/22.46} \\ \hline  
\end{tabular} 
}
\caption{The PSNR quantitative results of one-step and three-step inference for each initialization in four datasets.}
\label{tab:result_cros}
\end{table}

\begin{figure*}[!t]
\centering
\includegraphics[scale=0.29]{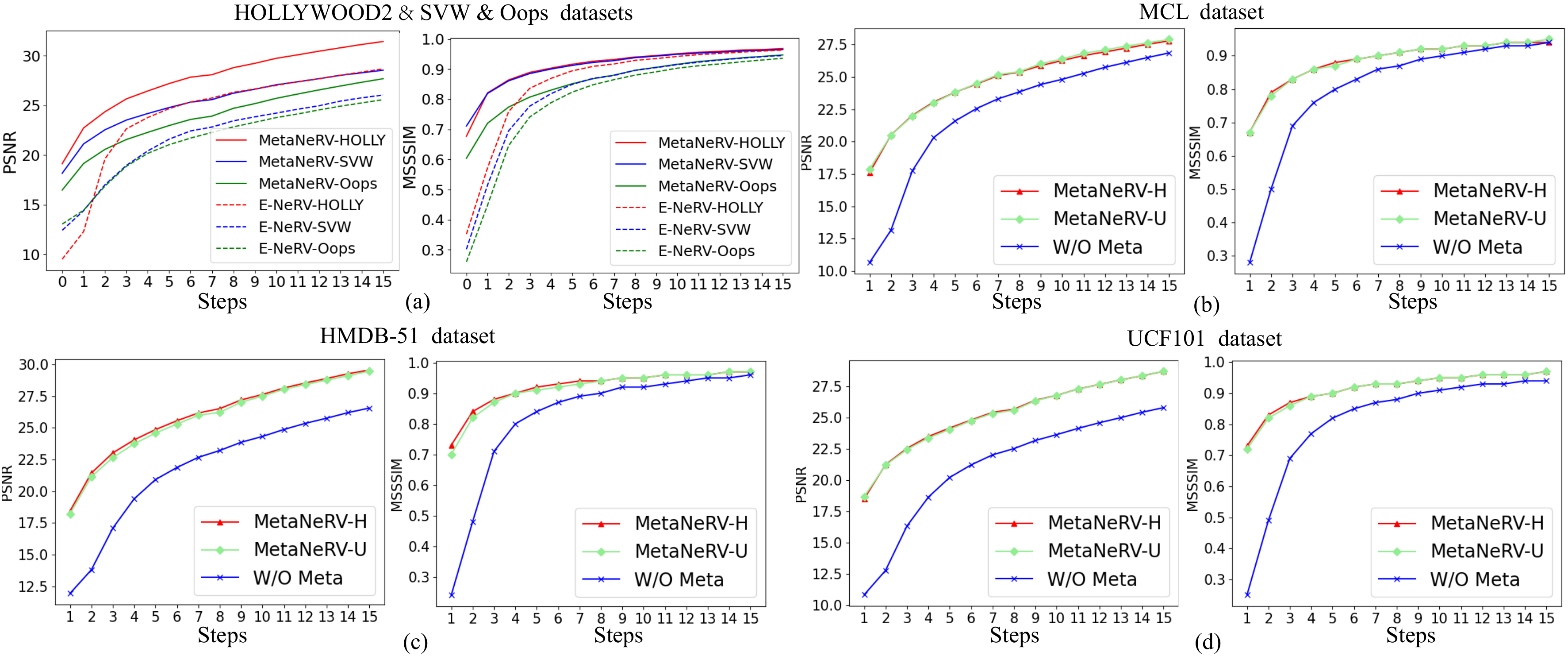}
\caption{
PSNR and MS-SSIM comparisons on steps. (a) Comparison of baselines E-NeRV, and our MetaNeRV on HOLLYWOOD2, SVW, and Oops datasets. (b)~(c)~(d) Comparison of random initialization, MetaNeRV-H and MetaNeRV-U on MCL$\_$JCV, HMDB-51, and UCF101 datasets. MetaNeRV-H denotes the optimal initialization representation derived from training on the HMDB-51 dataset. MetaNeRV-U denotes the optimal initialization representation derived from training on the UCF101 dataset. W/O Meta denotes E-NeRV(random initialization).
}
\label{fig:out_plot}
\end{figure*}

\begin{figure*}[!t]
\centering
\includegraphics[scale=0.34]{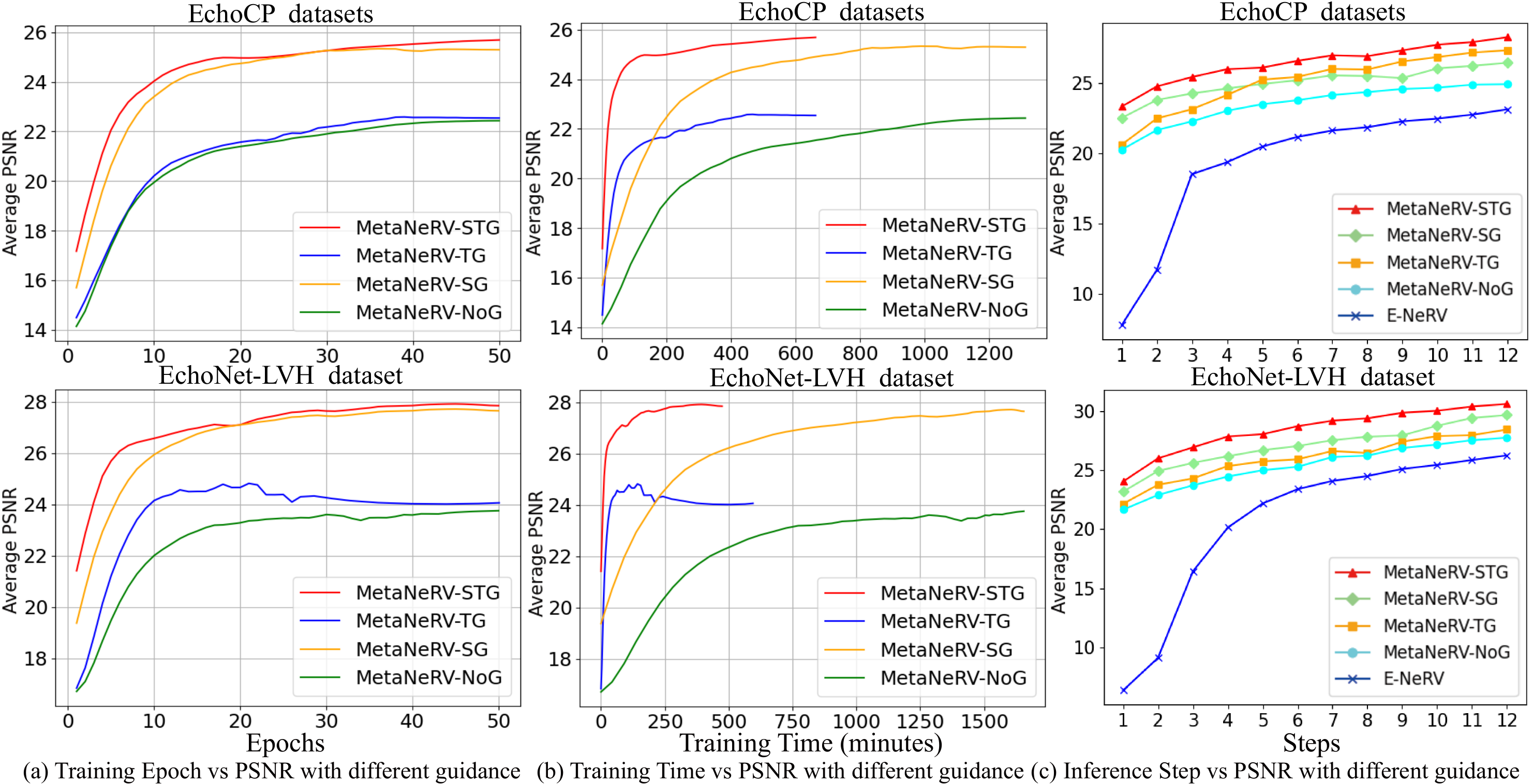}
\caption{
Ultrasound datasets comparison. 
(a) Training curves of Epochs, demonstrating improved model performance with spatial guidance.
(b) Training curves of Time, proving reduced training time and enhanced efficiency with temporal guidance.
(c) Inference performance of our method variants surpasses the baseline.
}
\label{fig:psnr and ssim 2}
\end{figure*}

\begin{figure}[!t]
\centering
\includegraphics[scale=0.5]{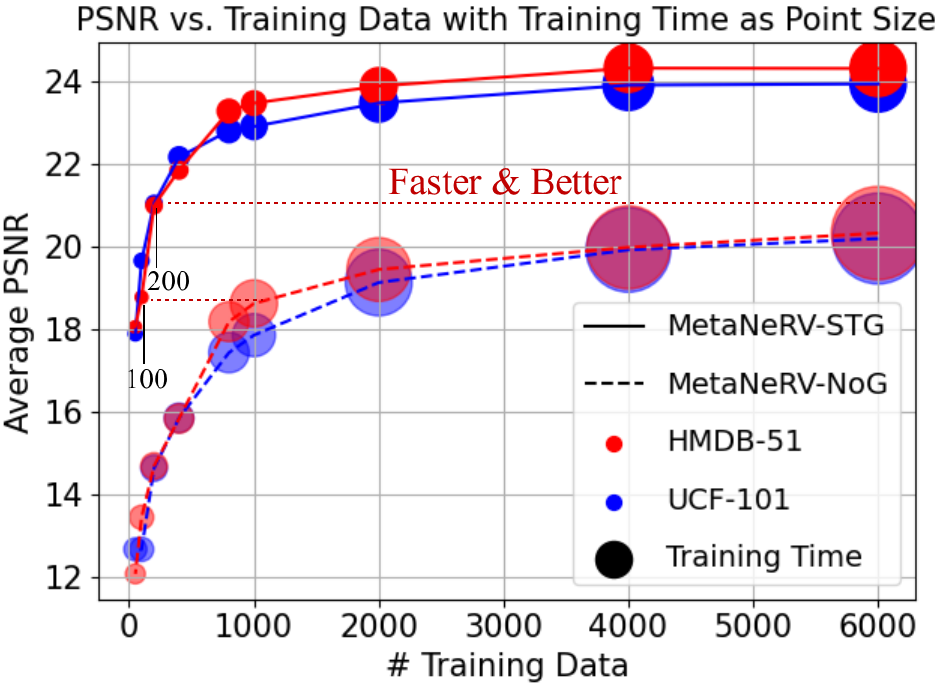}
\caption{
Training data scale vs average PSNR values with training time as point size on UCF101 and HMDB51 datasets, comparing MetaNeRV-NoG and MetaNeRV-STG. 
}
\label{fig:psnr and data}
\end{figure}

\begin{figure*}[!t]
\centering
\includegraphics[scale=0.24]{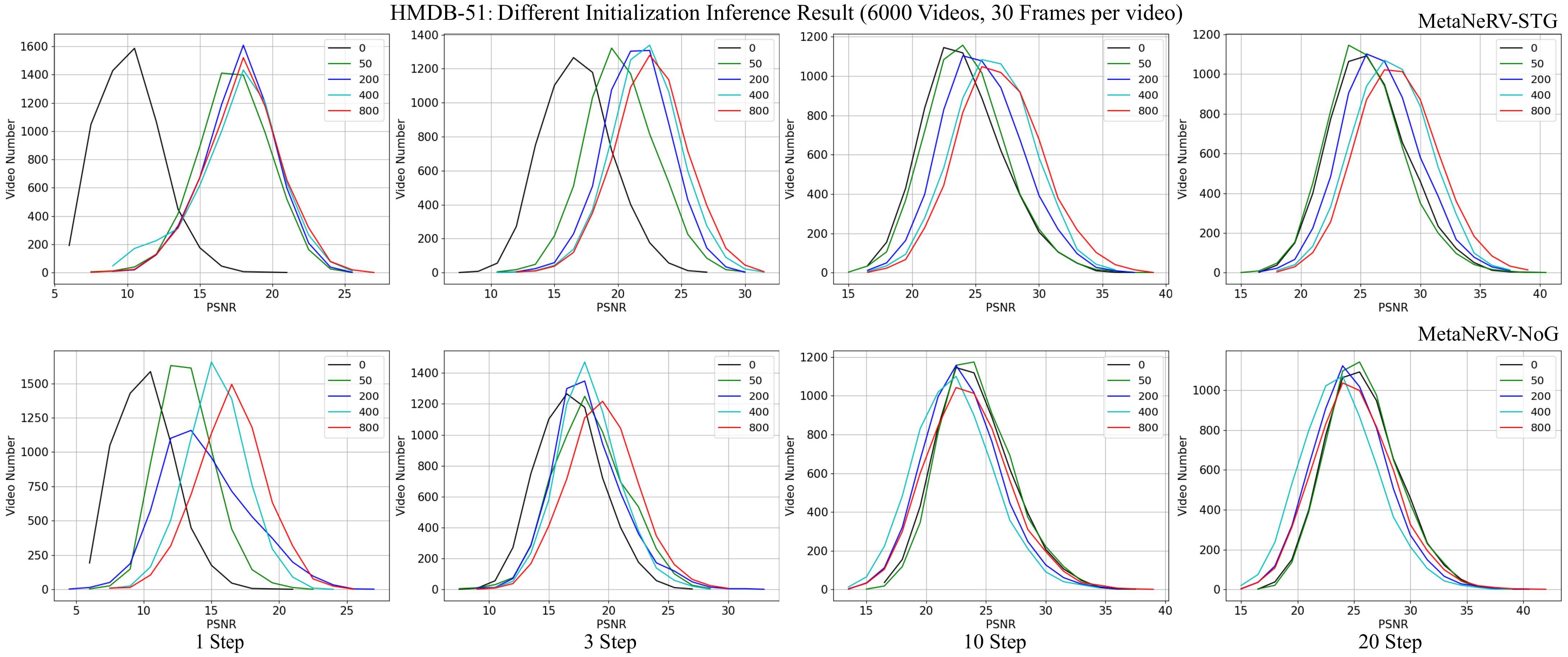}
\caption{
Different training data scale inference performance on the HMDB-51 dataset, video number is counted by different PSNR values, comparing MetaNeRV-NoG and MetaNeRV-STG.
}
\label{fig:num and psnr 1}
\end{figure*}

\begin{figure*}[!t]
\centering
\includegraphics[scale=0.24]{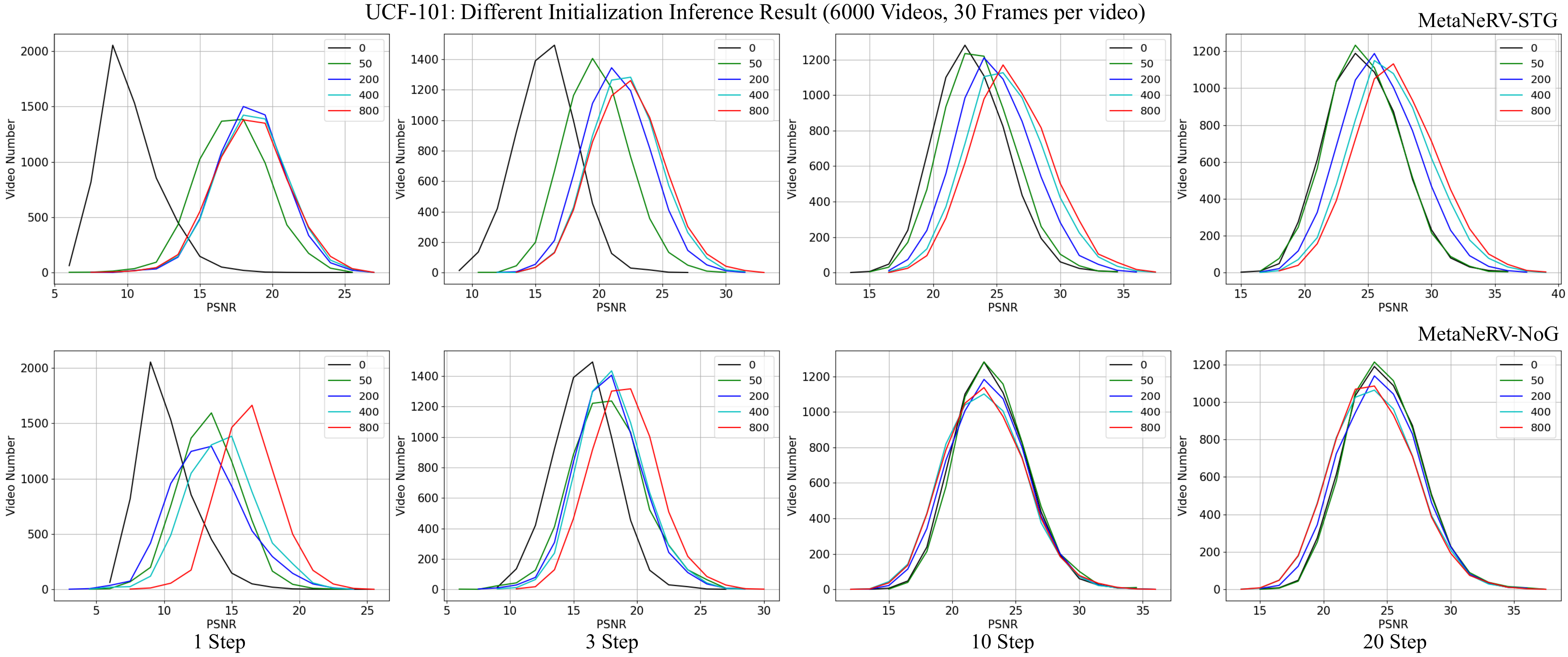}
\caption{
Different training data scale inference performance on the UCF-101 dataset, video number is counted by different PSNR values, comparing MetaNeRV-NoG and MetaNeRV-STG.
}
\label{fig:num and psnr 2}
\end{figure*}

\subsection{D. Additional Results}

\textbf{Impact of Dataset Scale on Meta-Initialization training and Inference Performance}. 
The dataset's scale significantly affects the representation capacity and generalization of meta-initialization. As the dataset expands, meta-initialization's representation capacity and generalization gradually enhance. Still, the computational cost also increases, resulting in longer training times for obtaining a superior meta-initialization.

Experimental results in figure~\ref{fig:psnr and data} show that a straightforward combination of meta-learning and the NeRV-base model (MetaNeRV-NoG) benefits from larger datasets. However, its performance, even with 6000 training videos, doesn't surpass that of our proposed method (MetaNeRV-STG) with just 200 videos. 
Our method, incorporating spatio-temporal guidance, improves with more training videos up to about 2000, after which performance stabilizes.

Our study demonstrates that adjusting the initialization parameters can markedly improve video representation effectiveness. Specifically, meta-initialization methods, with our proposed MetaNeRV-STG leading the way, consistently outperform random initialization across various reasoning steps. Notably, while both MetaNeRV-STG and MetaNeRV-NoG exhibit strong performance in the initial iterations, the effectiveness of MetaNeRV-NoG gradually diminishes as the number of iterations increases.

Figures~\ref{fig:num and psnr 1} and figures~\ref{fig:num and psnr 2} provide detailed insights into this conclusion. These figures present the statistical distributions of meta-initialization, obtained through training sets comprising different numbers of videos, across various inference steps on two datasets of 6000 videos. The videos are segmented and counted based on 1.5 PSNR value intervals. The curves in these figures are color-coded to indicate varying training video quantities, with 0 representing random initialization and 50 denoting a training set of 50 videos. From left to right, the distributions correspond to different inference steps, with the upper layer showcasing the MetaNeRV-STG method and the lower layer depicting the MetaNeRV-NoG method. The visual representations clearly illustrate the superiority of our meta-initialization techniques, particularly MetaNeRV-STG, in enhancing video representation effectiveness throughout the inference process.

\textbf{More OOD Quantitative results}. Note that our method significantly outperforms baseline methods on independent (a.k.a. OOD) test datasets even when trained on a single real-world dataset, demonstrating the robustness and generalizability of MetaNeRV.
The ablation studies on ultrasound datasets prove the effect of different guidance, as shown in Fig.~\ref{fig:psnr and ssim 2}.
As illustrated in Fig.\ref{fig:out_plot}(a), we utilized parameters trained on the HMDB-51 dataset as the optimal initialization parameters for direct inference on the three new datasets. 
The solid line represents our method, while the dashed line represents the E-NeRV method. 
The other three sub-figures in Fig.\ref{fig:out_plot} demonstrate the performance of our method on the OOD dataset. It can be observed that the optimal initialization parameters trained on the real-world dataset can be well transferred to other real-world datasets, indicating that our method is not affected by data distribution and can be quickly transferred to other datasets with just training on one real-world dataset.

\textbf{More OOD Qualitative results}. When adopting the model to specialized domains, we acknowledge that the optimal initial parameters should ideally be obtained from data with the same distribution. However, due to diverse samples in the training set, our model trained on real-world datasets exhibits good generalization capabilities compared to random initialization. See results in Tab.~\ref{tab:result_cros}. Parameters trained on general real-world datasets still generalize well to the ultrasound dataset, performing significantly better than random initialization. 
This highlights the effectiveness of our model's generalizability.

Additional PSNR result for the video representation task on the MCL$\_$JCV~\cite{wang2016mcl} dataset is summarized in Tab.~\ref{tab:psnr-values}. This comparison underscores the proficiency of our method in substantially decreasing convergence time.

\begin{figure*}[!t]
\centering
\includegraphics[scale=0.22]{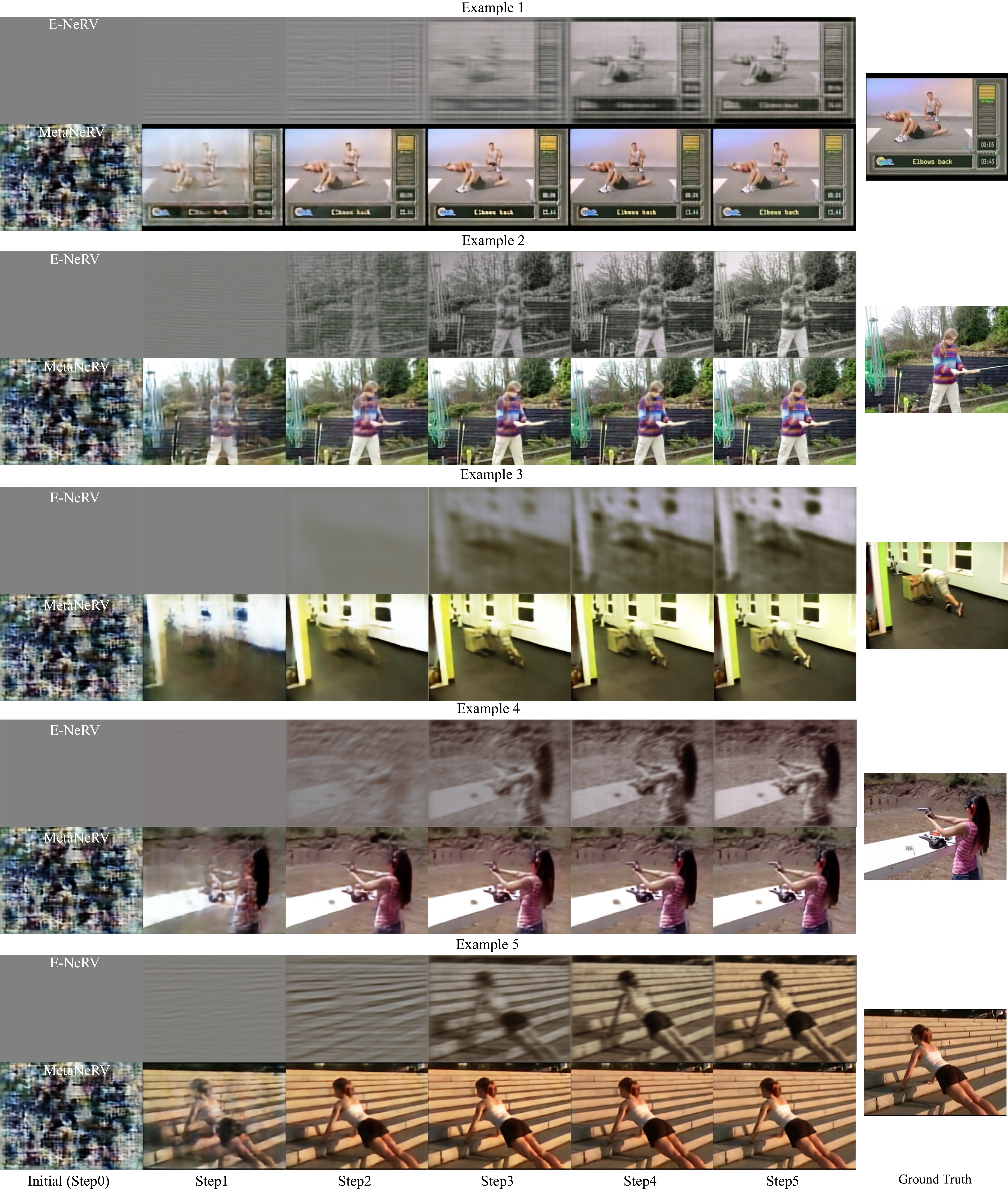}
\caption{The visualization of E-NeRV and MetaNeRV for different steps of the fitted HMDB-51 datasets videos. respectively, Our method yields impressive outcomes within just 5 steps of iterations.}
\label{fig:hmdb_s}
\end{figure*}

\begin{figure*}[!t]
\centering
\includegraphics[scale=0.22]{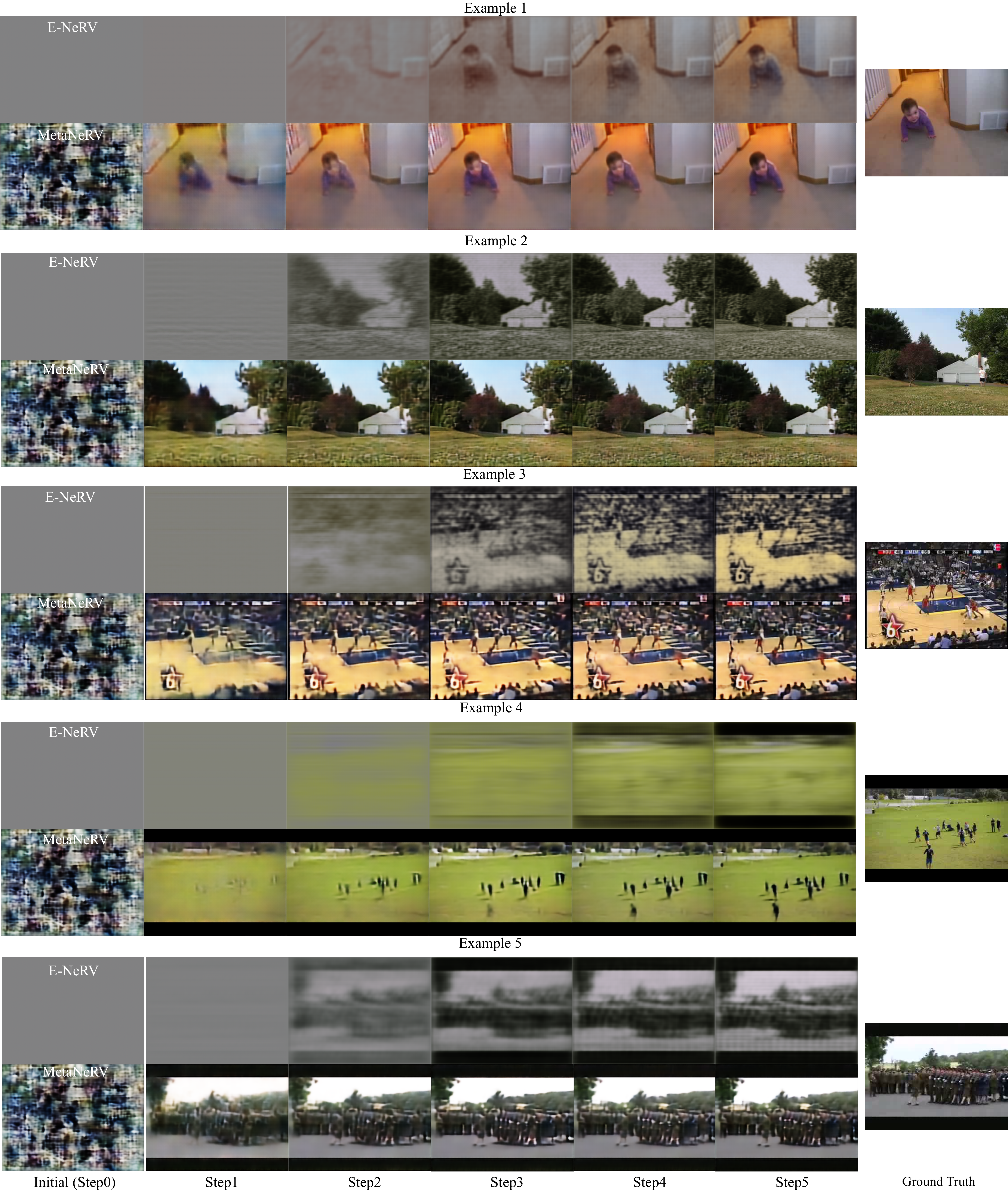}
\caption{The visualization of E-NeRV and MetaNeRV for different steps of the fitted UCF101 datasets videos. respectively, Our method yields impressive outcomes within just 5 steps of iterations.}
\label{fig:UCF_s}
\end{figure*}

\begin{figure*}[!t]
\centering
\includegraphics[scale=0.22]{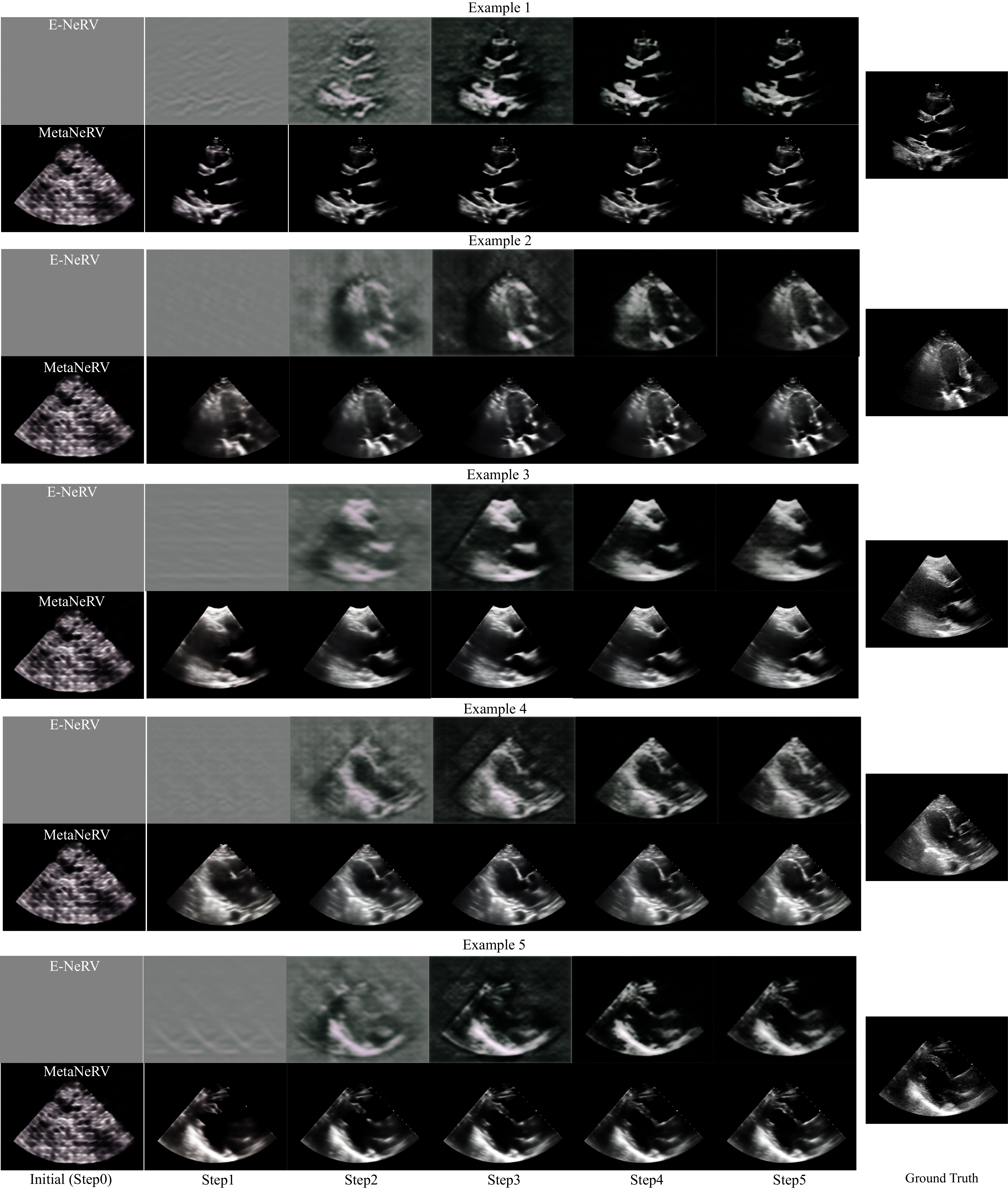}
\caption{The visualization of E-NeRV and MetaNeRV for different steps of the fitted EchoNet-LVH datasets videos. respectively, Our method yields impressive outcomes within just 5 steps of iterations.}
\label{fig:800_s}
\end{figure*}

\begin{figure*}[!t]
\centering
\includegraphics[scale=0.22]{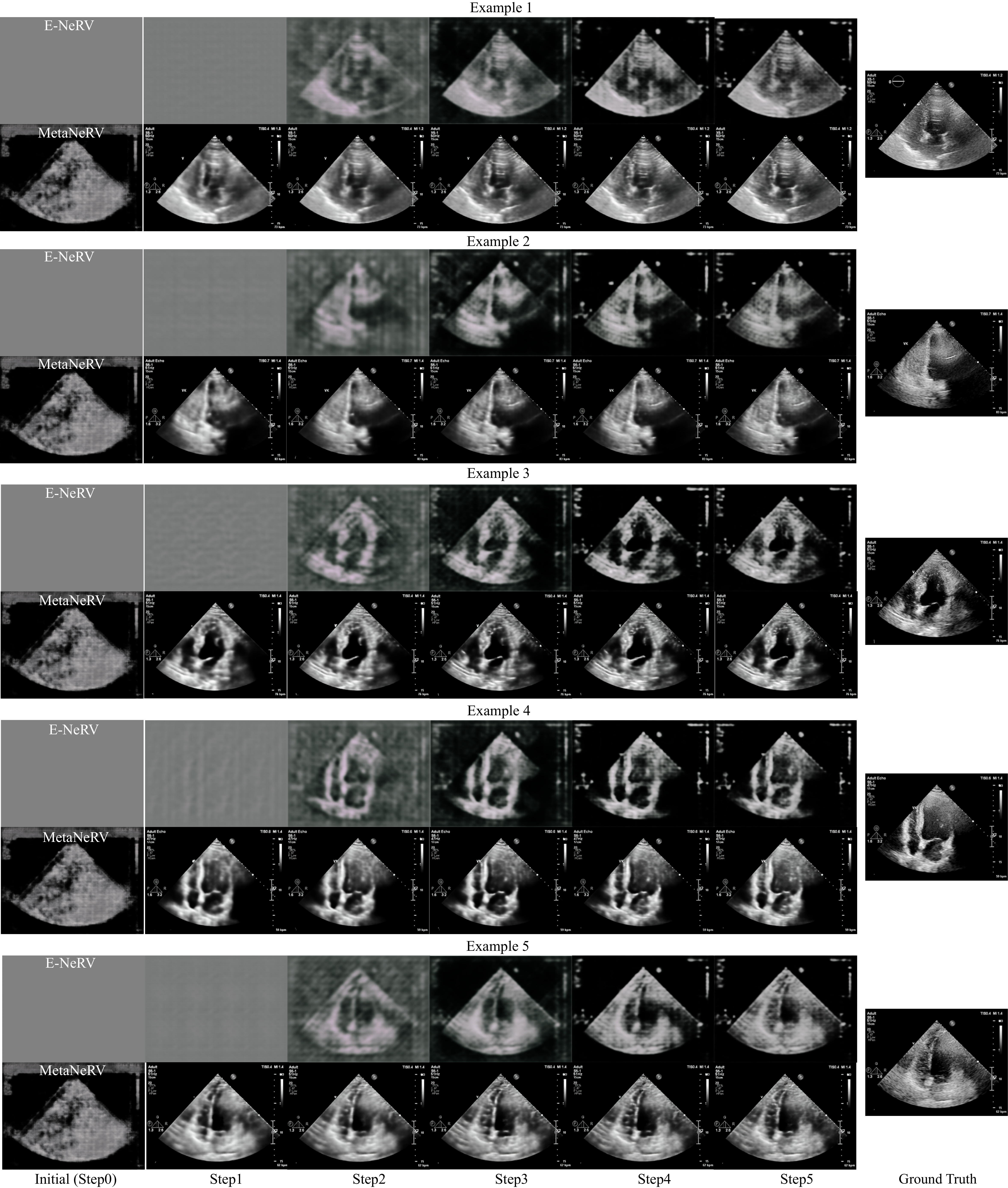}
\caption{The visualization of E-NeRV and MetaNeRV for different steps of the fitted EchoCP datasets videos. respectively, Our method yields impressive outcomes within just 5 steps of iterations.}
\label{fig:EchoCP_s}
\end{figure*}

\textbf{More visualization results}. Additional visual comparisons between the outputs from E-NeRV~\cite{li2022nerv} and MetaNeRV are given in Fig.~\ref{fig:hmdb_s}, Fig.~\ref{fig:UCF_s}, Fig.~\ref{fig:800_s}, Fig.~\ref{fig:EchoCP_s}. Despite having a few steps of iteration, the output from MetaNeRV is still noticeably better, with more detail from the original video frames preserved.

From Fig.~\ref{fig:hmdb_s} and Fig.~\ref{fig:UCF_s}, we observe that our approach can capture the approximate colors and shapes of the ground truth video in the first iteration by altering the initialization weights. 
In the second iteration step, our method has closely approximated the real video, reconstructing the primary cartoon characters and the background. Meanwhile, the shapes of the cartoon characters and background in the image reconstructed by E-NeRV are more pronounced.
In subsequent iteration steps, our method has achieved sufficient clarity to discern the finer details of the cartoon characters and the background. 

Our approach performs exceptionally well on medical datasets as observed in Fig.~\ref{fig:800_s} and Fig.~\ref{fig:EchoCP_s}. We hypothesize that the videos in medical ultrasound datasets are often collected using similar ultrasound equipment. Consequently, the initialization representation learned by our method already encapsulates the blurred sector shape and black background characteristic of medical ultrasound videos. Additionally, the monochromatic nature of ultrasound videos contributes significantly to the performance enhancement of our model. Our method can achieve representation results similar to the original video after just one iteration, and subsequent iterations primarily focus on refining the details within the video. In contrast, random initialization typically learns the black background within a few iterations but struggles to capture fine details, resulting in only a rough shape representation.

\begin{table*}[]  
\centering  
\caption{The quantitative results of different steps inference for two methods (E-NeRV/\textbf{MetaNeRV}) in MCL$\_$JCV datasets.}  
\label{tab:psnr-values}  
\begin{tabular}{cccccccc}  
\hline  
\textbf{Video ID} & \textbf{step1} & \textbf{step2} & \textbf{step3} & \textbf{step9} & \textbf{step15} & \textbf{step20} & \textbf{step25}\\ \hline  
videoSRC01 & 9.21/\textbf{24.85} & 11.54/\textbf{28.82} & 19.17/\textbf{30.4} & 29.09/\textbf{35.53} & 31.7/\textbf{37.81} & 32.75/\textbf{38.67} & 33.24/\textbf{39.07} \\
videoSRC02 & 12.16/\textbf{18.11} & 14.22/\textbf{19.55} & 17.07/\textbf{20.26} & 21.45/\textbf{22.92} & 22.8/\textbf{25.07} & 23.37/\textbf{26.19} & 23.59/\textbf{26.69} \\
videoSRC03 & 9.59/\textbf{21.56} & 12.15/\textbf{24.11} & 18.99/\textbf{25.14} & 24.93/\textbf{28.29} & 26.51/\textbf{30.19} & 27.21/\textbf{31.08} & 27.46/\textbf{31.5} \\
videoSRC04 & 11.78/\textbf{18.65} & 13.52/\textbf{20.29} & 16.65/\textbf{20.98} & 20.74/\textbf{22.51} & 21.69/\textbf{23.27} & 22.04/\textbf{23.63} & 22.16/\textbf{23.78} \\
videoSRC05 & 14.75/\textbf{16.49} & 15.77/\textbf{18.54} & 17.41/\textbf{19.46} & 20.81/\textbf{22.38} & 22.3/\textbf{24.21} & 22.88/\textbf{25.21} & 23.12/\textbf{25.54} \\
videoSRC06 & 9.37/\textbf{9.57} & 10.85/\textbf{13.12} & 15.44/\textbf{15.62} & 25.29/\textbf{28.37} & 31.93/\textbf{33.38} & 35.54/\textbf{37.98} & 37.75/\textbf{39.24} \\
videoSRC07 & 12.98/\textbf{19.99} & 15.23/\textbf{22.69} & 19.42/\textbf{23.87} & 23.21/\textbf{27.05} & 24.96/\textbf{28.93} & 25.66/\textbf{29.83} & 25.94/\textbf{30.23} \\
videoSRC08 & 7.86/\textbf{17.24} & 10.28/\textbf{22.35} & 18.16/\textbf{23.71} & 23.89/\textbf{27.4} & 25.24/\textbf{29.08} & 25.93/\textbf{29.61} & 26.21/\textbf{29.91} \\
videoSRC09 & 11.84/\textbf{15.76} & 13.62/\textbf{17.44} & 14.83/\textbf{18.38} & 18.92/\textbf{21.77} & 20.96/\textbf{23.81} & 21.77/\textbf{24.65} & 22.08/\textbf{25.02} \\
videoSRC10 & 10.59/\textbf{13.13} & 11.59/\textbf{13.64} & 12.33/\textbf{13.83} & 14.28/\textbf{14.8} & 14.98/\textbf{15.52} & 15.49/\textbf{16.05} & 15.72/\textbf{16.37} \\
videoSRC11 & 10.74/\textbf{16.87} & 12.45/\textbf{19.32} & 15.96/\textbf{20.41} & 20.61/\textbf{23.27} & 22.25/\textbf{24.97} & 22.9/\textbf{26.0} & 23.18/\textbf{26.41} \\
videoSRC12 & 11.76/\textbf{19.43} & 13.55/\textbf{22.33} & 17.42/\textbf{23.87} & 23.17/\textbf{28.4} & 25.84/\textbf{31.3} & 27.15/\textbf{32.49} & 27.61/\textbf{32.93} \\
videoSRC13 & 14.28/\textbf{17.41} & 15.24/\textbf{19.57} & 16.78/\textbf{20.44} & 21.45/\textbf{24.6} & 24.41/\textbf{28.48} & 25.91/\textbf{29.95} & 26.49/\textbf{30.6} \\
videoSRC14 & 10.6/\textbf{17.09} & 12.82/\textbf{19.14} & 15.77/\textbf{20.0} & 19.22/\textbf{22.61} & 20.88/\textbf{24.37} & 21.61/\textbf{25.09} & 21.89/\textbf{25.42} \\
videoSRC15 & 9.61/\textbf{20.08} & 12.24/\textbf{22.46} & 17.84/\textbf{23.48} & 22.49/\textbf{25.78} & 24.41/\textbf{27.34} & 25.09/\textbf{27.99} & 25.33/\textbf{28.31} \\
videoSRC16 & 7.66/\textbf{15.66} & 9.61/\textbf{22.61} & 16.15/\textbf{26.38} & 27.29/\textbf{30.65} & 29.02/\textbf{32.2} & 29.69/\textbf{32.69} & 29.93/\textbf{33.04} \\
videoSRC17 & 10.44/\textbf{20.83} & 13.43/\textbf{22.49} & 19.52/\textbf{23.41} & 22.86/\textbf{25.61} & 24.03/\textbf{26.85} & 24.54/\textbf{27.47} & 24.76/\textbf{27.77} \\
videoSRC18 & 11.33/\textbf{17.14} & 13.64/\textbf{19.01} & 16.12/\textbf{20.04} & 19.89/\textbf{22.53} & 21.59/\textbf{24.16} & 22.15/\textbf{24.99} & 22.37/\textbf{25.31} \\
videoSRC19 & 11.73/\textbf{18.32} & 13.43/\textbf{19.52} & 16.85/\textbf{20.23} & 20.25/\textbf{22.61} & 21.38/\textbf{24.46} & 22.01/\textbf{25.5} & 22.26/\textbf{25.84} \\
videoSRC20 & 14.11/\textbf{16.9} & 15.05/\textbf{19.11} & 16.33/\textbf{20.07} & 20.66/\textbf{22.6} & 22.66/\textbf{24.42} & 23.48/\textbf{25.25} & 23.8/\textbf{25.71} \\
videoSRC21 & 8.6/\textbf{20.5} & 10.91/\textbf{25.43} & 19.33/\textbf{27.39} & 26.65/\textbf{32.72} & 29.04/\textbf{35.45} & 30.78/\textbf{36.5} & 31.34/\textbf{36.92} \\
videoSRC22 & 10.95/\textbf{18.22} & 13.85/\textbf{19.7} & 17.53/\textbf{20.56} & 20.61/\textbf{22.63} & 21.89/\textbf{24.4} & 22.38/\textbf{25.22} & 22.59/\textbf{25.56} \\
videoSRC23 & 11.56/\textbf{17.1} & 13.34/\textbf{20.49} & 16.53/\textbf{21.88} & 22.06/\textbf{25.92} & 23.82/\textbf{27.85} & 24.6/\textbf{28.73} & 24.9/\textbf{29.1} \\
videoSRC24 & 8.4/\textbf{17.71} & 10.75/\textbf{20.26} & 16.71/\textbf{21.13} & 20.47/\textbf{23.76} & 21.89/\textbf{25.5} & 22.78/\textbf{26.26} & 23.08/\textbf{26.67} \\
videoSRC25 & 9.32/\textbf{15.07} & 10.02/\textbf{17.15} & 12.19/\textbf{18.44} & 18.68/\textbf{21.43} & 20.41/\textbf{22.68} & 21.08/\textbf{23.31} & 21.35/\textbf{23.55} \\
videoSRC26 & 10.89/\textbf{17.84} & 12.96/\textbf{19.6} & 15.62/\textbf{20.78} & 20.38/\textbf{24.39} & 22.0/\textbf{26.46} & 22.91/\textbf{27.71} & 23.25/\textbf{28.17} \\
videoSRC27 & 10.79/\textbf{19.92} & 12.92/\textbf{22.84} & 17.1/\textbf{24.1} & 23.14/\textbf{27.09} & 25.39/\textbf{29.12} & 26.46/\textbf{30.16} & 26.82/\textbf{30.48} \\
videoSRC28 & 7.48/\textbf{15.29} & 9.58/\textbf{20.67} & 15.48/\textbf{23.99} & 25.7/\textbf{30.11} & 28.52/\textbf{32.67} & 29.66/\textbf{33.29} & 30.07/\textbf{33.64} \\
videoSRC29 & 7.54/\textbf{14.12} & 9.45/\textbf{21.89} & 15.43/\textbf{25.26} & 32.86/\textbf{38.7} & 35.19/\textbf{41.08} & 36.23/\textbf{41.8} & 36.66/\textbf{42.23} \\
videoSRC30 & 13.05/\textbf{15.75} & 15.38/\textbf{21.28} & 18.11/\textbf{24.49} & 26.27/\textbf{31.12} & 29.34/\textbf{33.76} & 30.23/\textbf{34.78} & 30.56/\textbf{35.26} \\ \hline
means & 10.69/\textbf{17.55} & 12.64/\textbf{20.51} & 16.74/\textbf{21.93} & 22.91/\textbf{25.91} & 24.73/\textbf{27.96} & 25.54/\textbf{28.94} & 25.85/\textbf{29.34} \\
\hline  
\end{tabular}  
\end{table*}

